\documentclass[letter, 10pt, conference]{ieeeconf}      

\IEEEoverridecommandlockouts                              

\usepackage{graphicx} 
\usepackage{amsmath} 
\usepackage{subcaption} 
\usepackage{tikz} 
\usepackage{pgfplots} 
\usepackage{booktabs} 
\usepackage{multirow} 
\usepackage{url}
\usepackage{hyperref}

\usepackage{cleveref}
\usepackage[acronyms]{glossaries}

\newacronym{fmcw}{FMCW}{Frequency-Modulated Continuous-Wave}
\newacronym{oord}{OORD}{Oxford Offroad RobotCar Dataset}
\newacronym{sar}{SAR}{Synthetic-Aperture Radar}

\crefname{section}{Sec.}{Secs.}
\crefname{table}{Tab.}{Tabs.}
\crefname{figure}{Fig.}{Figs.}
\crefname{equation}{Eq.}{Eqs.}

\title{\LARGE \bf
Beyond a Shadow of a Doubt: Close Proximity Geometry Reconstruction Using FMCW Radar Shadow Effects
}

\author{
Felix de Trogoff du Boisguezennec$^{1,2}$, Benjamin Ramtoula$^1$, Daniele De Martini$^1$ \\
$^1$ Oxford Robotics Institute, University of Oxford, UK \\
$^2$ ETH Zurich, Switzerland
\thanks{The work was supported by the EPSRC Programme Grant ``From Sensing to Collaboration'' (EP/V000748/1).
Corresponding author: \texttt{daniele@robots.ox.ac.uk}.}
}

\begin{document}


\maketitle
\thispagestyle{empty}
\pagestyle{empty}

\begin{abstract}
Reliable perception in adverse conditions remains challenging for autonomous systems, as cameras and LiDAR degrade in poor lighting or weather.
Millimetre-wave FMCW radar is robust to such conditions, but its elevation collapse limits geometric reasoning. We observe that vehicle chassis occlude radar rays and form a distinctive geometric shadow, and its consistency can enable us to infer useful information about objects whose returns intersect this shadow. Motivated by this observation, we propose a method to recover the 3D, in-plane inclination of nearby slender vertical objects from this cue. The object inclination is retrieved without assumptions about the wider scene, but through an analytical,  closed-form mapping between its radar return boundaries and the opening angle. Validation in simulation and experimentation on a Navtech CTS350-X radar shows that inclinations can be estimated under practical conditions, with segmentation of the object in the radar scan emerging as the main bottleneck. This work highlights chassis shadows as a novel geometric cue, extending the role of 2D rotating radar beyond localisation and toward 3D scene reconstruction.
\end{abstract}

\section{Introduction}
Autonomous robots require reliable perception for navigation and safety, yet cameras and LiDAR degrade under fog, rain, snow, or poor lighting \cite{BijelicCVPR2020_SeeingThroughFog,HahnerICCV2021_FogSimLiDAR,HahnerCVPR2022_SnowLiDAR}.
Radar provides robustness in such conditions \cite{PaekNeurIPS2022_KRadar,CaesarCVPR2020_nuScenes}, but its coarse resolution, sparse returns, and elevation integration have limited its use for geometry estimation of the surrounding scene. Existing radar perception methods often rely on heuristics, priors, learned 3D neural representations, or additional modalities~\cite{AllandSPM2019_InterferenceSurvey,KrausIROS2021_GhostObjects, Borts2024_RadarFields, radarsplat}.

\begin{figure}[t]
    \centering
    \begin{subfigure}{0.45\textwidth}
        \centering
        \includegraphics[width=0.65\textwidth]{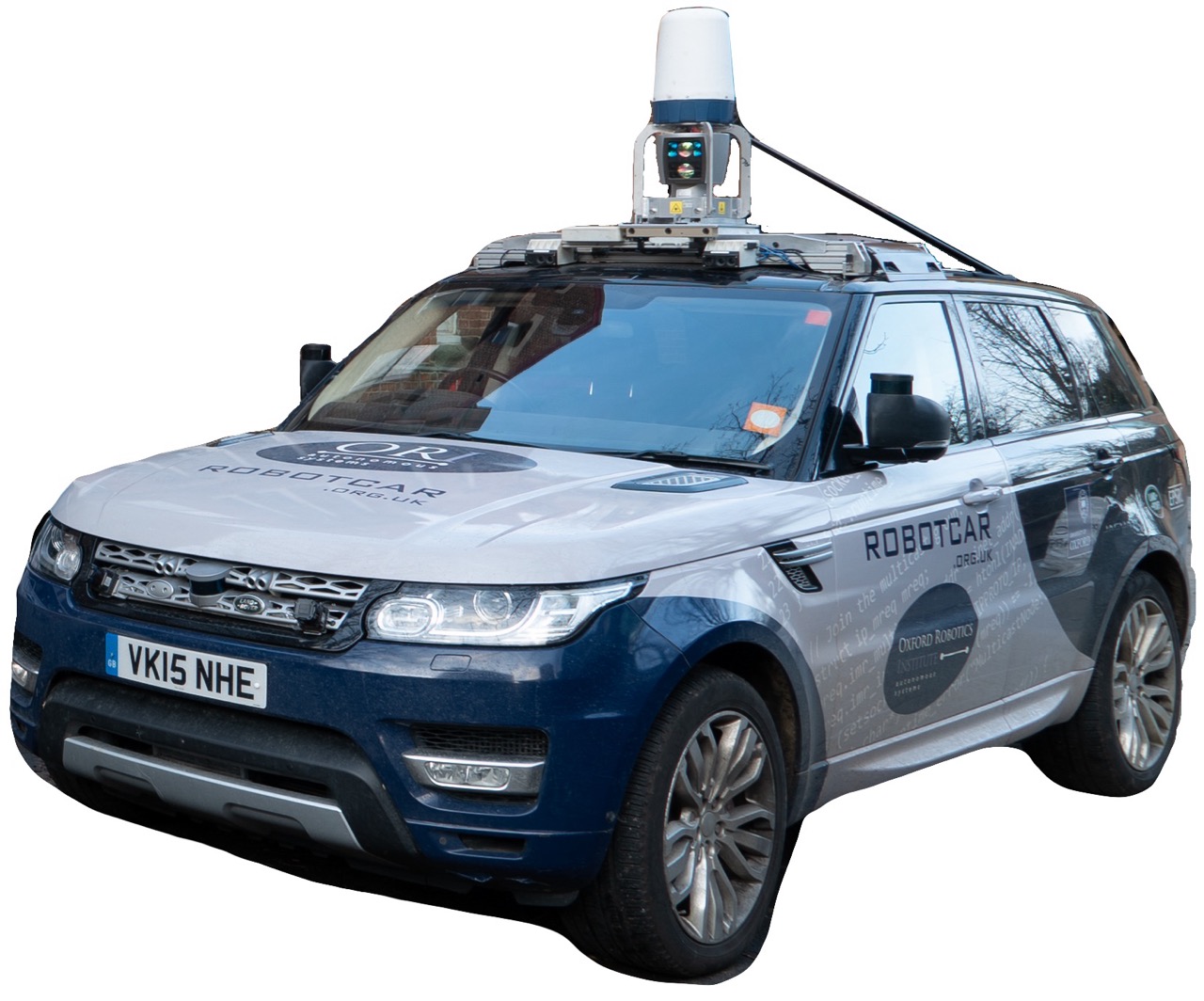}
        \caption{\label{fig:car}}
    \end{subfigure}
    
    \begin{subfigure}{0.45\textwidth}
        \centering
        \includegraphics[width=0.67\textwidth]{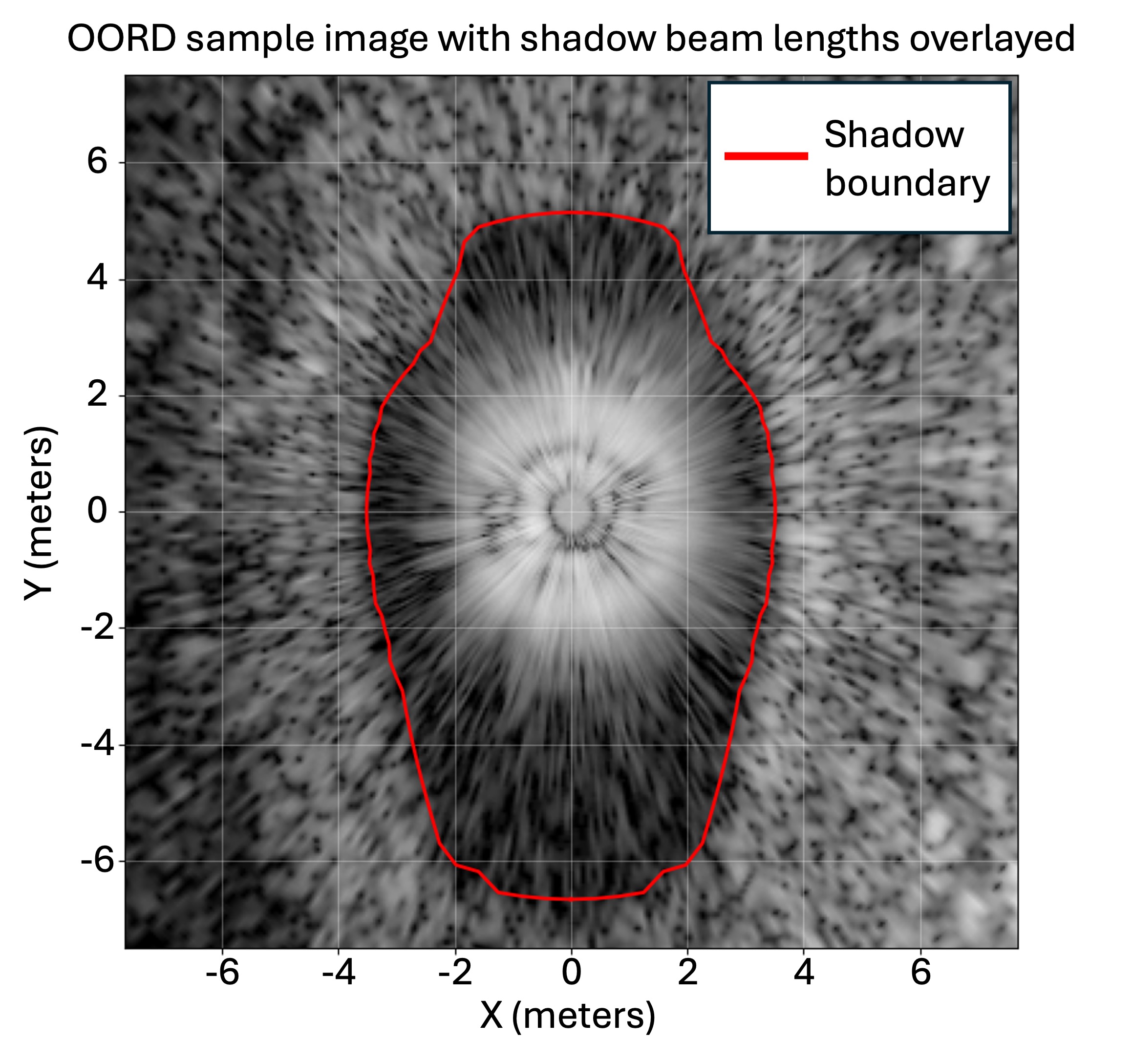}
        \caption{\label{fig:shadow}}
    \end{subfigure}
    \caption{(a) Roof-mounted Navtech CTS350-X on the \acrshort{oord} \cite{Gadd2024_OORD} platform.
    (b) When stationary on a flat surface, the hull of the car casts a radar shadow on the ground. This shadow aligns perfectly with the overlaid red line depicting the predicted shadow boundary from a CAD model. In this work, we use knowledge of this shadow's properties to infer surrounding objects' inclination.}
    \vspace{-15pt}
\end{figure}

\Gls{fmcw} scanning radars rotate about their vertical axis while sensing the environment continuously through the transmission and reception of frequency-modulated radio waves.
While rotating, the sensor inspects one angular portion -- azimuth -- at a time.
The radar transmits radio waves and receives its reflected power signal, which is a function of reflectivity, size and orientation of objects in the environment.
\cref{fig:car} shows the setup used in the \gls{oord} \cite{Gadd2024_OORD} -- and adopted in this work -- where the radar is mounted on the top of the car.

\begin{figure}[t]
    \centering
    \includegraphics[width=0.45\textwidth]{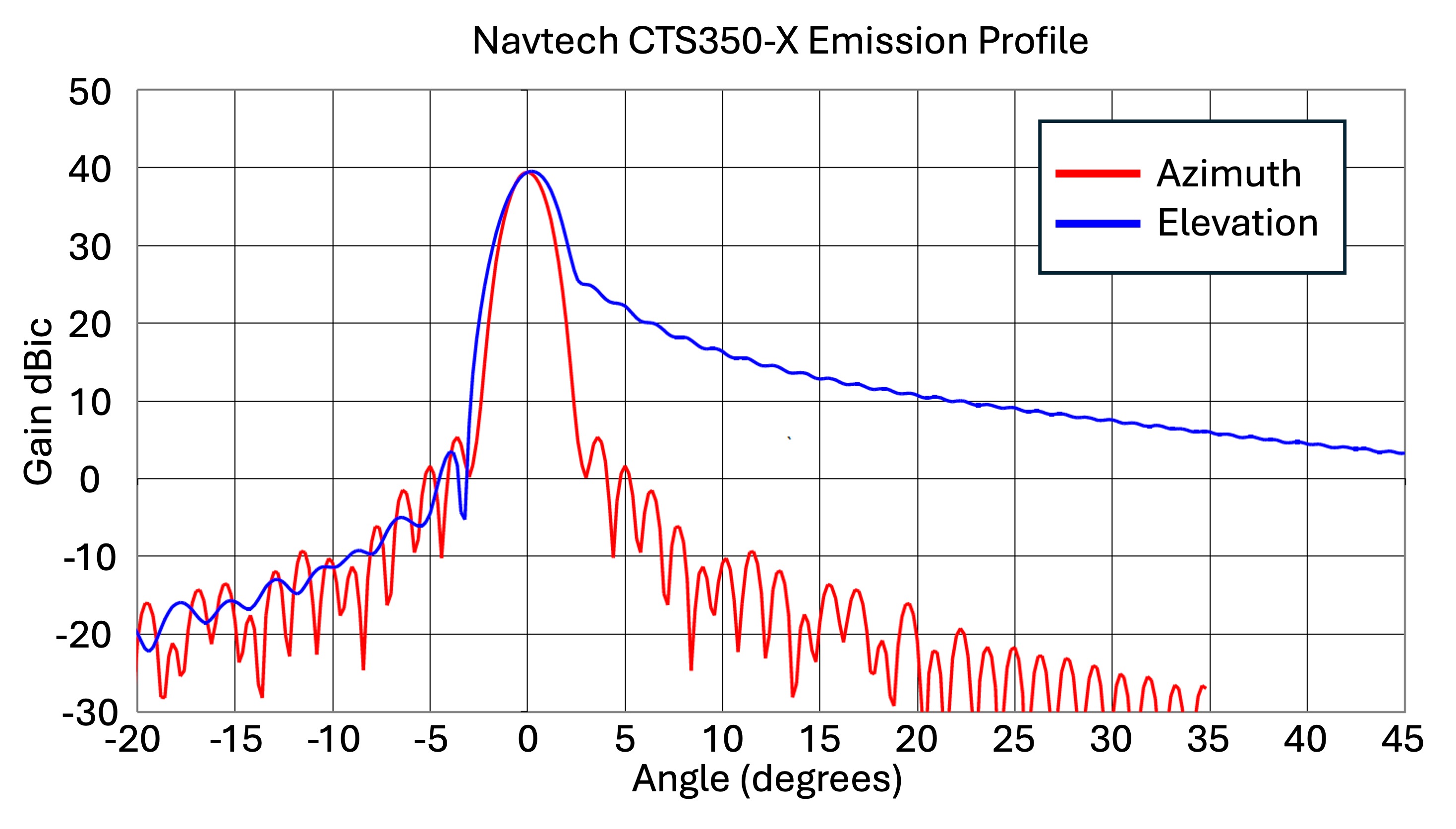}
    \caption{Emission profile showing the spread of the radar returns in the azimuth direction (red) and elevation (blue). This profile shows how objects outside the theoretical direction of emission influence the radar signal. A strength threshold of 5 dBic defines the usable range, which in elevation extends from $3^\circ$ above to $35^\circ$ below horizontal.}
    \label{fig:oord_emission}
    \vspace{-15pt}
\end{figure}

Unlike a LiDAR scanner, the nature of the radio waves leads to a reflected signal that originates from a volumetric slice rather than a punctual region in space -- theoretically, on the horizontal plane on the sampled azimuth ray direction; consequently, objects present in the space close to the inspected azimuth ray will affect the returned signal.
\cref{fig:oord_emission} shows how the return signal is a function of the angle deviation from the reported azimuth, both in azimuth direction and, more critically, in the elevation from the horizontal plane, i.e. elements below the horizontal plane affect the sensor readings.

\cref{fig:shadow} shows how the vehicle chassis naturally occludes radar beams, creating a proximal shadow zone.
So far, this effect has been seen as a limitation of radar from the manufacturers, who have been trying to focus the radar wave more and more on the theoretical sensing point.
Here, instead, \textit{we leverage this property} and introduce a proof-of-concept method for estimating the inclination of nearby slender vertical objects using only 2D rotating FMCW radar. Because real-world environments often violate idealised flatness assumptions we here rely on and available driving radar datasets lack ground truth for such subtle geometric parameters, our scope is strictly geared towards formulating and validating this effect in a highly controlled setting to determine its applicability in more complex tasks.

By analysing the shadow geometry in conjunction with the vehicle's CAD model, we derive a closed-form mapping from the measurable shape of the object in the radar image to its inclination on the radar plane.
The method requires no assumptions about the wider environment, making it suitable for unstructured terrain.
It requires only simple calculations and works without calibration in new environments, making it an ideal solution for onboard vehicles to measure the inclination of their immediate surroundings, such as poles or trees.

The approach is validated both in physics-based simulation (RadaRays \cite{Mock2023_RadaRays}) and with a Navtech CTS350-X radar, demonstrating feasibility and identifying key error sources. Our contributions are:
\begin{enumerate}
    \item A validation of the presence of previously overlooked geometrical cues about surroundings through radar shadows.
    \item A geometric formulation linking radar shadows and object inclination.
    \item Validation in simulation and real data, showing practical recoverability of inclinations.
\end{enumerate}
Although our present work addresses inclination, our results reveal that radar scans contain valuable information that has so far been overlooked. This opens the door to future methods that, for example, estimate ground plane geometry.

\section{Related Work}
\label{sec:related}

Cameras and LiDARs degrade in poor weather, motivating radar as a robust alternative.
Benchmarks and surveys highlight radar's resilience and explore fusion strategies with optical sensors \cite{BijelicCVPR2020_SeeingThroughFog,Zhang2023_AdverseWeatherSurvey,Zhou2022_DeepRadarPerceptionSurvey}.

Spinning \gls{fmcw} radars, such as the Navtech CTS350-X, support odometry, localisation, and place recognition, with datasets spanning urban and off-road conditions \cite{BarnesICRA2020_ORRD,Gadd2024_OORD}.
Methods range from classical estimation to learning-based odometry and radar-only SLAM \cite{CenICRA2018_PreciseRadarEgomotion,BarnesCoRL2020_MaskingByMoving,Burnett2021_RadarOdometryRSS,Qiao2025_RadarTeachRepeat,cfear,tang2023point,radarslam}.
While some works, either on purpose or by using learned feature extractors, have leveraged the radar elevation returns \cite{cen2019,burnett2021radar,tang2021self}, all still exploit statistical stability in radar images, but do not use platform-induced occlusions as geometric cues.
For example, \cite{cen2019} exploits the elevation returns to detect keypoints for odometry on the ground in a very unstructured, off-road environment; however, this effect does not seem to be intentional, but a byproduct of the radar beam’s vertical spread and ground interactions, rather than a deliberate use of platform-induced occlusions as a source of geometric information.

Most radar-based reconstruction relies on 4D MIMO sensors or neural scene representations that use per-return elevation and Doppler to recover 3D structure \cite{radarsplat,PaekNeurIPS2022_KRadar, Borts2024_RadarFields} A recent relevant approach also leveraged known radar properties and sensor motion to resolve elevation of stationary objects~\cite{hou2023automotive}. 
These approaches are complementary but differ from mechanically rotating 2D scanners with elevation collapse.

In \gls{sar} and remote sensing, shadows have been used to recover shape and building height \cite{Yang2020_VideoSAR_ShadowGMTI,Sun2022_SAR_BuildingHeight,Xie2021_SAR_ShadowHeight}. However, these are far-field imaging regimes. To our knowledge, no prior work leverages vehicle chassis shadows in 2D rotating radar for near-field geometry.

Prior work demonstrates radar's robustness, the use of spinning radars for localisation, and shadows in remote sensing. Our contribution is the first to exploit chassis-induced shadows in 2D rotating \gls{fmcw} radar to recover object inclination without environmental assumptions.

\section{Method}
\label{sec:method}

\subsection{Validating the source of radar shadows}
\begin{figure}[t]
    \centering
    \includegraphics[width=0.85\linewidth]{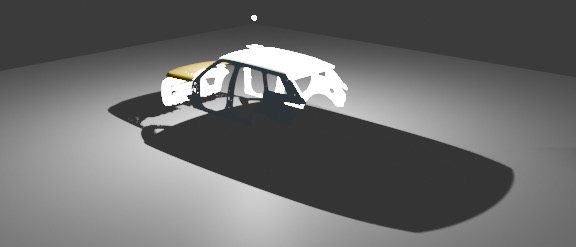}
    \caption{Blender recreation of the \acrshort{oord} setup, used for the analysis of this work. It is clear how the CAD model produces a shadow of the radar sensor on the ground. Such a shadow can be seen over real data in \cref{fig:shadow}.}
    \label{fig:blender_images}
    \vspace{-15pt}
\end{figure}

In datasets that employ a rotating \gls{fmcw}  radar mounted on a vehicle, a dark halo consistently appears around the sensor origin.
This effect can be seen in \cref{fig:shadow} in the \gls{oord} \cite{Gadd2024_OORD}.
We hypothesise that this effect arises from the vehicle chassis occluding part of the radar beam.
To validate this geometric occlusion hypothesis, we recreated the \gls{oord} setup in Blender using a CAD model of the Range Rover vehicle and a simple point source emitter at the radar pose (\cref{fig:blender_images}). While this point light acts as a purely geometric proxy to compute line-of-sight occlusion and does not model the complex interactions of radio waves, this simplified experiment already shows a good validation of our hypothesis.
The resulting ground-plane shadow footprint matches the symmetric dark zone observed in dataset averages, confirming that the effect is a stable chassis-induced occlusion rather than an environmental artefact -- \cref{fig:shadow} shows the CAD footprint on a single radar scan.

Our key insight is that \textit{this chassis shadow is a constant geometric feature} directly tied to platform dimensions -- i.e. it does not depend on external effects.
It defines an azimuth-dependent opening angle $\theta$ that can be exploited as a reference for recovering object orientation without assumptions about the wider environment.

\subsection{Geometric model: relating scan contents to inclination of surrounding objects}
\begin{figure}[t]
    \centering
    \includegraphics[width=0.9\linewidth]{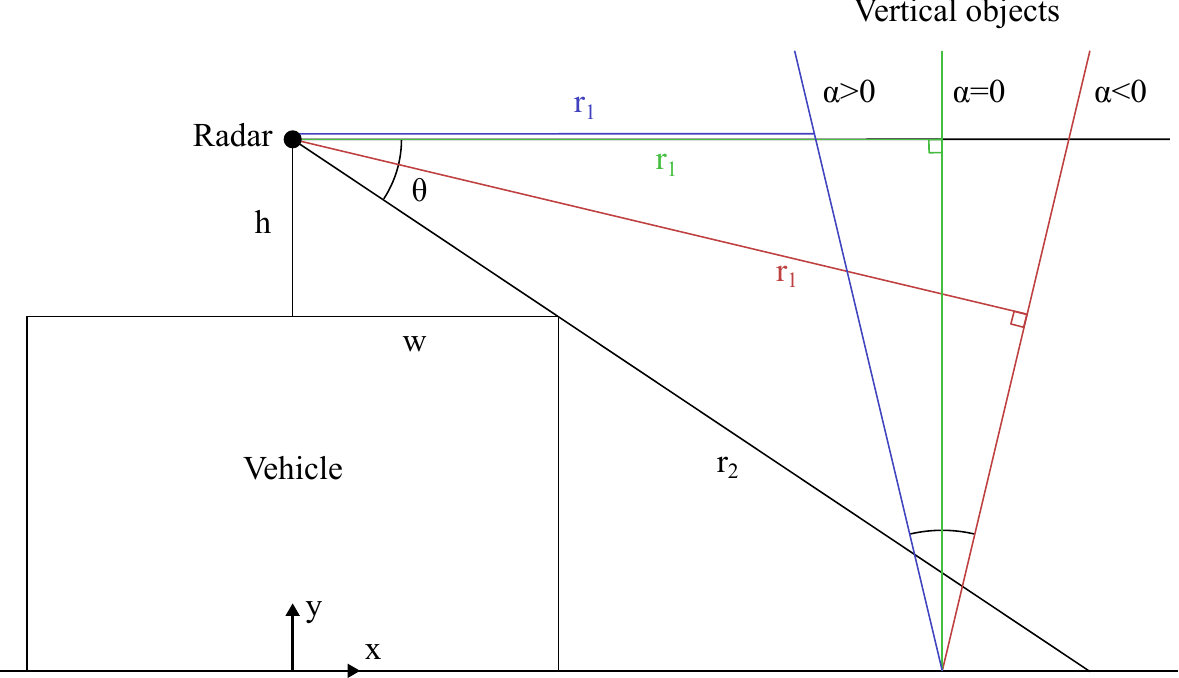}
    \caption{Shadow geometry: chassis opening $\theta$, measured bounds $(r_1,r_2)$, and target object inclination $\alpha$.}
    \label{fig:model_scheme}
    \vspace{-10pt}
\end{figure}

We first model the set-up of the vehicle and the radar, as illustrated in \cref{fig:model_scheme}.
Let the radar be mounted at height $h$ on the vehicle, and the chassis half-width be $w$. The chassis edge defines the opening angle
\[
\theta=\arctan\!\left(\tfrac{h}{w}\right), \qquad \theta<35^\circ,
\]
where $35^\circ$  corresponds to the observed elevation spread in the radar's emission profile (\cref{fig:oord_emission}), leading to a requirement in geometry for the beam to clear the chassis. We also model the top boundary of the emission profile to be horizontal.

Now, let us assume the vehicle has a target object in its surroundings, which we are interested in detecting using radar scans.
Specifically, we consider a slender vertical target located within the radar shadow. The slenderness assumption allows us to model the object as lying along a single azimuth of the radar scan. To make the inclination observable, the object must span from the shadow boundary at its base to the horizontal emission boundary at its top. These two geometric constraints are sufficient to uniquely determine the pole’s in-plane inclination.

On a given azimuth, such a slender vertical target inside the shadow produces a contiguous return. We denote $r_1$: the inner radial bound (first detection), $r_2$: the outer bound (last detection), and $\alpha$: the in-plane inclination of the target relative to vertical. Thus, $\theta$ is fixed by vehicle geometry, $(r_1,r_2)$ are directly observed from radar scans after segmentation of the target object, and $\alpha$ is the unknown to be recovered. Two cases arise depending on whether the pole leans towards the radar ($\alpha\ge0$, closer to sensor) or away from it ($\alpha<0$, further from sensor), which will be uniquely distinguished algebraically based on the bounds.

\noindent\textbf{Case 1: $\alpha \ge 0$.} Here $r_1$ lies on the horizontal boundary and $r_2$ on the shadow boundary. By applying the law of sines and cosine sum formulae to the triangle formed by the sensor origin and the bounds of the visible pole segment, we obtain:
\begin{equation}
\frac{r_2}{r_1}=\frac{\cos\alpha}{\cos(\theta+\alpha)}, 
\quad \alpha\in\big[0,\tfrac{\pi}{2}-\theta\big),
\end{equation}
which gives
\begin{equation}
\alpha=\arctan\!\left(\frac{(\tfrac{r_2}{r_1})\cos\theta-1}{(\tfrac{r_2}{r_1})\sin\theta}\right).
\end{equation}

\noindent\textbf{Case 2: $\alpha < 0$.} Now $r_1$ -- being the shortest distance from the radar -- slides along the pole, with $r_2$ remaining on the shadow boundary. Similarly applying the law of sines yields:

\begin{equation}
\frac{r_2}{r_1}=\frac{1}{\cos(\theta+\alpha)}, 
\quad \alpha\in\big[-\tfrac{\theta}{2},0\big),
\end{equation}
so that
\begin{equation}
\alpha=\arccos\!\left(\tfrac{r_1}{r_2}\right)-\theta.
\end{equation}

We note that there is a lower bound for $\alpha$ of $-\theta/2$, beyond which $r_1$ and $r_2$ become symmetric and introduce an ambiguity that requires further information -- e.g. from odometry estimation -- to resolve.

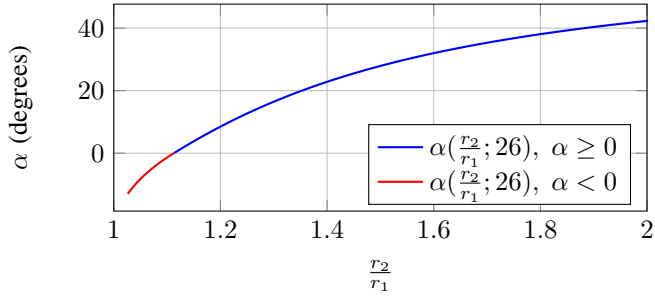
\begin{figure}[t]
\centering
\begin{tikzpicture}
\begin{axis}[
    xlabel={$\tfrac{r_2}{r_1}$},
    ylabel={$\alpha$ (degrees)},
    xmin=1, xmax=2,
    samples=200,
    grid=both,
    width=\linewidth,
    height=0.5\linewidth,
    legend pos=south east,
]

\pgfmathsetmacro{\theta}{26}

\addplot[blue, thick, domain={1/cos(\theta)}:5] 
    ({x}, {atan((x*cos(\theta)-1)/(x*sin(\theta)))});
\addlegendentry{$\alpha(\tfrac{r_2}{r_1};\theta), \;\alpha\ge0$}

\addplot[red, thick, domain={1/cos(\theta/2)}:{1/cos(\theta)}] 
    ({x}, {acos(1/x)-\theta});
\addlegendentry{$\alpha(\tfrac{r_2}{r_1};\theta), \;\alpha<0$}

\end{axis}
\end{tikzpicture}
\caption{Full inclination angle $\alpha$ vs. $\tfrac{r_2}{r_1}$ for $\theta = 26^\circ$ as from the CAD model of \gls{oord}.}
\label{fig:full_alpha_plot}
\vspace{-10pt}
\end{figure}

\noindent\textbf{Full model.} These equations give us a complete mapping allowing us to predict inclination $\alpha$ of a surrounding object using known fixed geometry of the vehicle $\theta$ and radar scan information $r_1$ and $r_2$:
\begin{equation}
\alpha\!\left(\rho;\theta\right)=
\begin{cases}
\arctan\!\left(\tfrac{\rho\cos\theta-1}{\rho\sin\theta}\right) \\
\hspace{4em}\text{for }\rho\ge 1/\cos\theta, \\[0.6em]
\arccos\!\bigl(1/\rho\bigr)-\theta \\
\hspace{4em}\text{for }\rho\in[1/\cos(\tfrac{\theta}{2}),\,1/\cos\theta).
\end{cases}
\label{eq:full_model}
\end{equation}
where $\rho = \tfrac{r_2}{r_1}$.
The two branches meet continuously at $\alpha=0$.
The function is shown in \cref{fig:full_alpha_plot}.
This formulation establishes a closed-form relationship between measurable radar quantities $(r_1,r_2)$, the known opening $\theta$, and the inclination $\alpha$ of a slender vertical target object.

\subsection{Processing radar scans}
The previous developments rely on extracting $r_1$ and $r_2$, the inner and outer radial bounds of the object, from scans. In practice this entails two steps: (i) segmentation to isolate the target return and (ii) boundary extraction with corrections for elongation.

\noindent\textbf{Segmentation.}
In simulation, target boundaries are well defined and $r_1,r_2$ are obtained by fitting a bounding box to the target return on the relevant azimuth.
For real data, we operate on a static scene and build a temporal consistency mask by intersecting $n$ consecutive frames, retaining only pixels that are equal in all $n$; this suppresses transient clutter and stabilises the target silhouette. We use $n=16$, which corresponds to 4 seconds of measurements.

\noindent\textbf{Range–elongation model.}
Real scans exhibit radial smearing: strong echoes start at the true range and smear outward, inflating the outer boundary. We model this with an additive elongation bias \(d\) on the observed outer range, i.e., we set the corrected outer boundary as \(r_2 = r_2^{\text{obs}} - d\) while retaining \(r_1 = r_1^{\text{obs}}\). Within our operating range, \(d\) is treated as approximately constant (estimated as described in Sec.~\ref{sec:Results}) and applied unchanged at test time.

Following this process, we are able to obtain $\theta$ from the known geometry, and $r_1$ and $r_2$ through our processing of radar scans, which enables us to estimate inclination through the relationships described in \cref{eq:full_model}.

\section{Experimental Setup}
\label{sec:Experimental setup}
Our evaluations focus on validating our approach's precision when predicting target objects' inclination.
We first test our work in simulation, allowing us to validate our approach in ideal conditions and its robustness to diverse configurations, such as different emission models or target diameters.
We then conduct real-world experiments to validate applicability, including sensor noise and reflections.

\subsection{Simulation Experimental Setup}

\noindent\textbf{Simulation Platform.} Radar simulators vary significantly in fidelity. 
Robotics tools like Gazebo\footnote{\url{gazebosim.org/}} 
typically provide generic sensor models with additive noise but omit detailed effects such as shadowing. Full-wave solvers such as CST Studio Suite\footnote{\url{3ds.com/products-services/simulia/products/cst-studio-suite/}} and Remcom XFdtd\footnote{\url{remcom.com/xfdtd-3d-em-simulation-software}} model electromagnetic propagation accurately but are computationally expensive. Automotive simulators such as CARLA \cite{Dosovitskiy2017_CARLA} support forward-looking MIMO radar models but do not handle mechanically rotating sensors. By contrast, RadaRays \cite{Mock2023_RadaRays} offers hardware-accelerated ray tracing, occlusion, multipath, and shadow simulation for rotating \gls{fmcw} radar and integrates into ROS/Gazebo. For these reasons, we proceeded with RadaRays for our experiments. Noise was disabled in experiments to isolate geometric effects.

\begin{figure}[t]
    \centering
    \begin{subfigure}{0.25\textwidth}
        \includegraphics[height=0.68\textwidth]{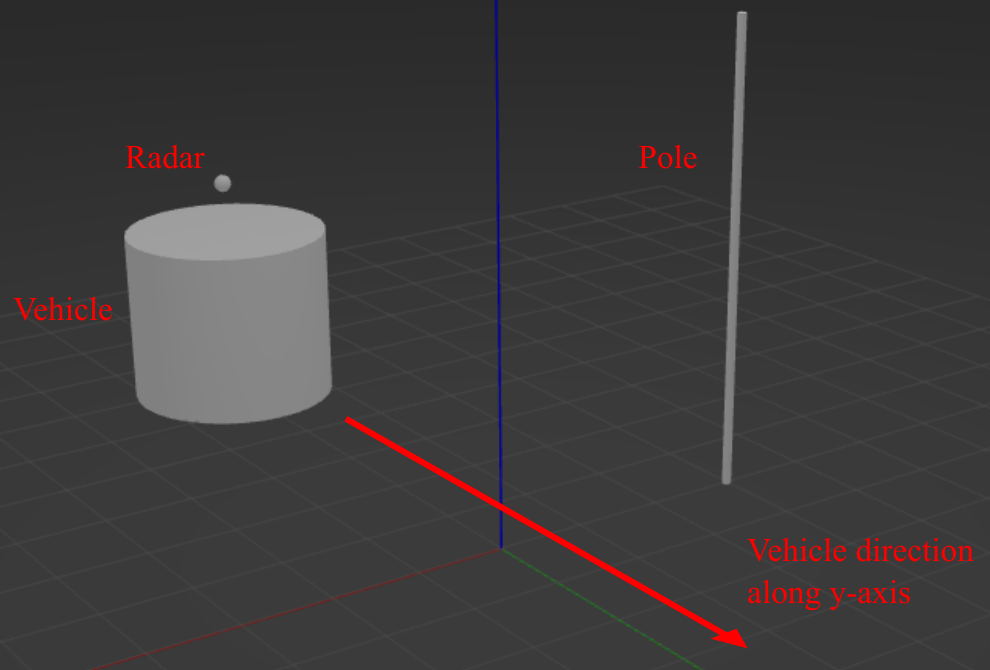}
        \caption{\label{fig:cylinder}}
    \end{subfigure}
    \begin{subfigure}{0.2\textwidth}
        \includegraphics[height=0.85\textwidth]{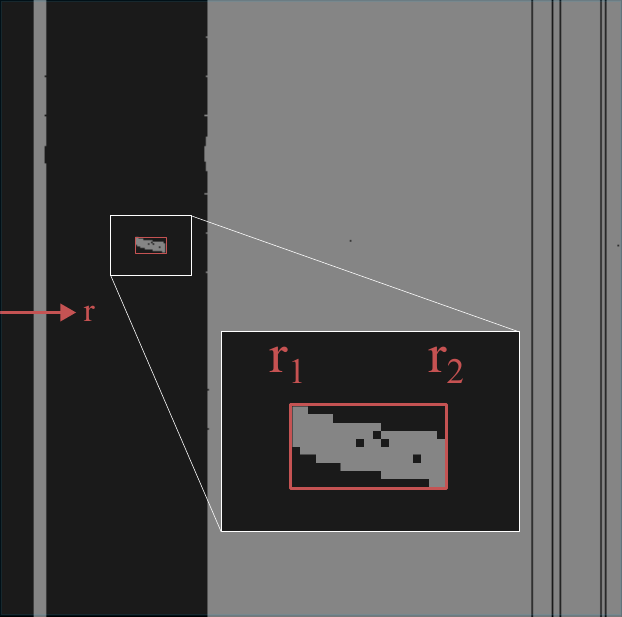}
        \caption{\label{fig:bounding_box}}
    \end{subfigure}
    \caption{(a) Simulation setup in Gazebo and (b) corresponding polar radar data with segmentation bounding box.}
    \label{fig:sim_setup_and_polar}
    \vspace{-10pt}
\end{figure}

\noindent\textbf{Experimental Design.} A simplified cylindrical chassis (\cref{fig:cylinder}) was used to produce a constant conical shadow across all azimuths, avoiding variations from detailed vehicle contours. The cylinder radius and radar mounting height were chosen such that $\theta=26^\circ$, matching the real-world setup. The vehicle advanced at constant velocity while the pole was positioned so that its base intersected the shadow cone as the vehicle passed. $r_1$ and $r_2$ were extracted by fitting bounding boxes around the radar return (\cref{fig:bounding_box}). Simulation parameters were varied to assess robustness: resolution (perfect vs.\ real sensor), emission model (ideal vs.\ measured from \cref{fig:oord_emission}), and pole diameter (1, 10, 20 cm). Four conditions were tested, combining these parameters, with poles tilted at $-10^\circ$, $0^\circ$, and $+10^\circ$.

\noindent\textbf{Analytical ground-truth function.}
Let the radar move along $(0,y,h)$, and let the pole pass through $(d,0,h)$ with true tilt $\psi$ (zenith angle within its inclination plane) and azimuth $\phi$ relative to the vehicle's forward axis. Then the measurable bounds on the relevant azimuth are
\[
r_1=\sqrt{y^2+d^2}, \qquad
\]
\[
r_2=\sqrt{(d-h\tan\psi\cos\phi)^2+(y+h\tan\psi\sin\phi)^2+h^2}.
\]
We define the ground-truth inclination function used for evaluation as
\begin{equation}
\alpha_{\mathrm{gt}}(y;d,\psi,\phi)=\arctan\!\left(\frac{\sqrt{r_2^2-h^2}-r_1}{h}\right).
\label{eq:alpha_gt}
\end{equation}

The reconstructed angle $\alpha_{\mathrm{gt}}$ varies with the vehicle motion $y$ and coincides with the pole’s true tilt only when the radar trajectory lies in the pole’s inclination plane. Away from that plane, the recovered value corresponds to an equivalent “rotated” pole configuration. Importantly, the mapping $(\psi,\phi,d)\mapsto \alpha_{\mathrm{gt}}(y)$ is injective: a measured curve uniquely determines the pole’s tilt, azimuth, and lateral offset. 

\begin{figure}[b]
    \centering
    \includegraphics[width=0.9\columnwidth]{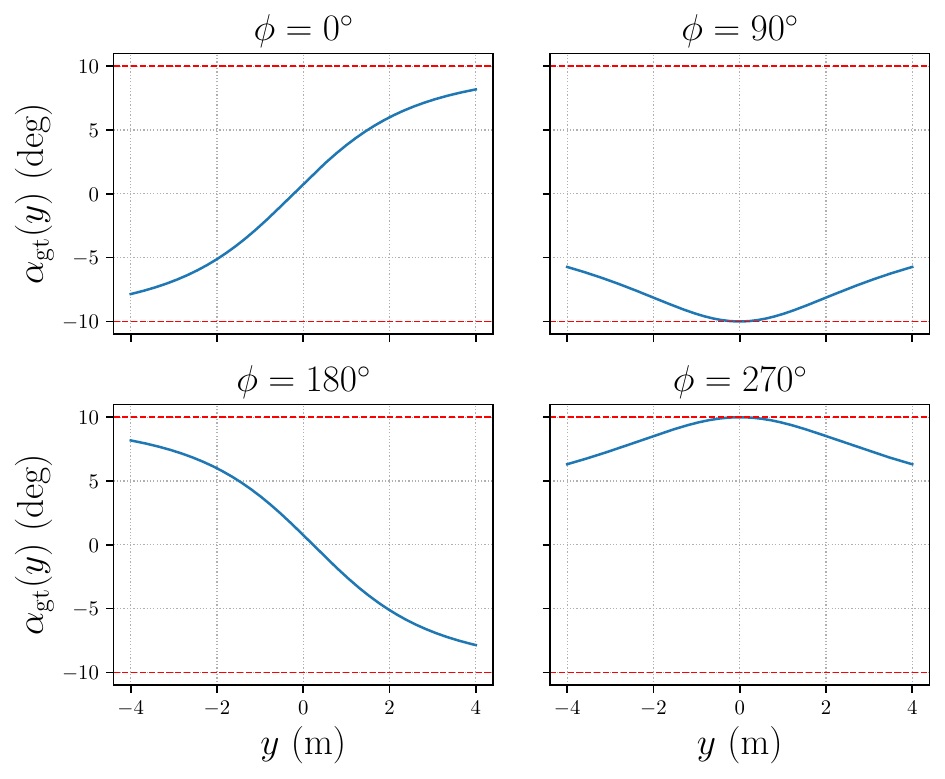}
    \caption{Family of $\alpha(y)$ curves for all azimuth orientations $\phi \in [0,360^\circ)$, with $\psi=10^\circ$, $d=-3\,\mathrm{m}$, $h=2.580\,\mathrm{m}$}
    \label{fig:alpha_graphs_circle}
\end{figure}

To illustrate the behaviour of $\alpha_{\mathrm{gt}}(y;d,\psi,\phi)$, 
\cref{fig:alpha_graphs_circle} shows its variation across azimuths $\phi\in[0,360^\circ)$ for $\psi=10^\circ$. 

\subsection{Real-World Experimental Setup}
\begin{figure}[!htbp]
    \centering
    \includegraphics[height=0.7\linewidth]{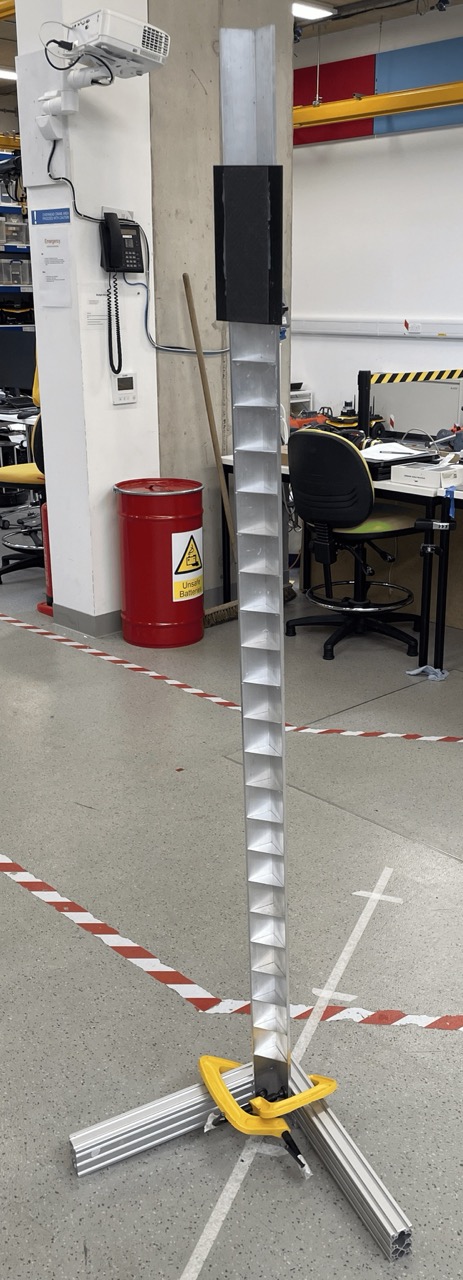}\hfill
    \includegraphics[height=0.7\linewidth]{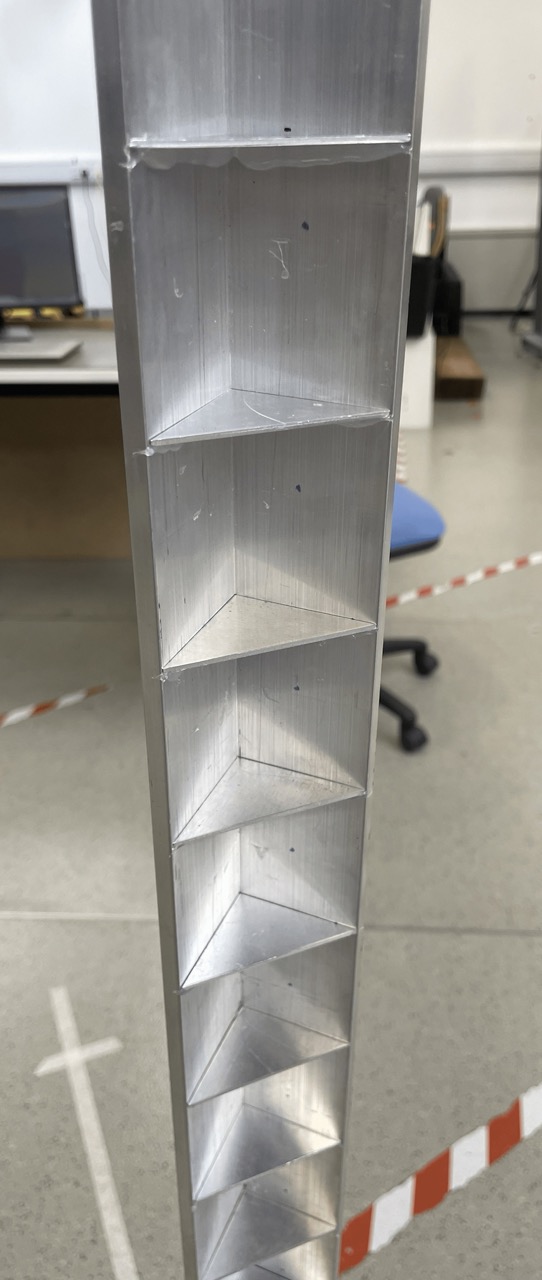}\hfill
    \includegraphics[height=0.7\linewidth]{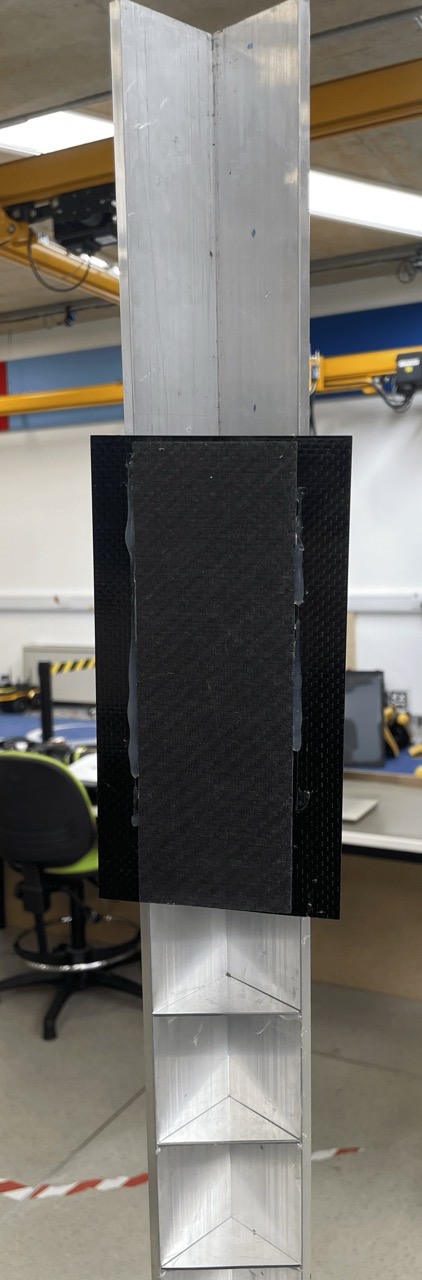}

    \par\medskip
    \includegraphics[height=0.37\linewidth]{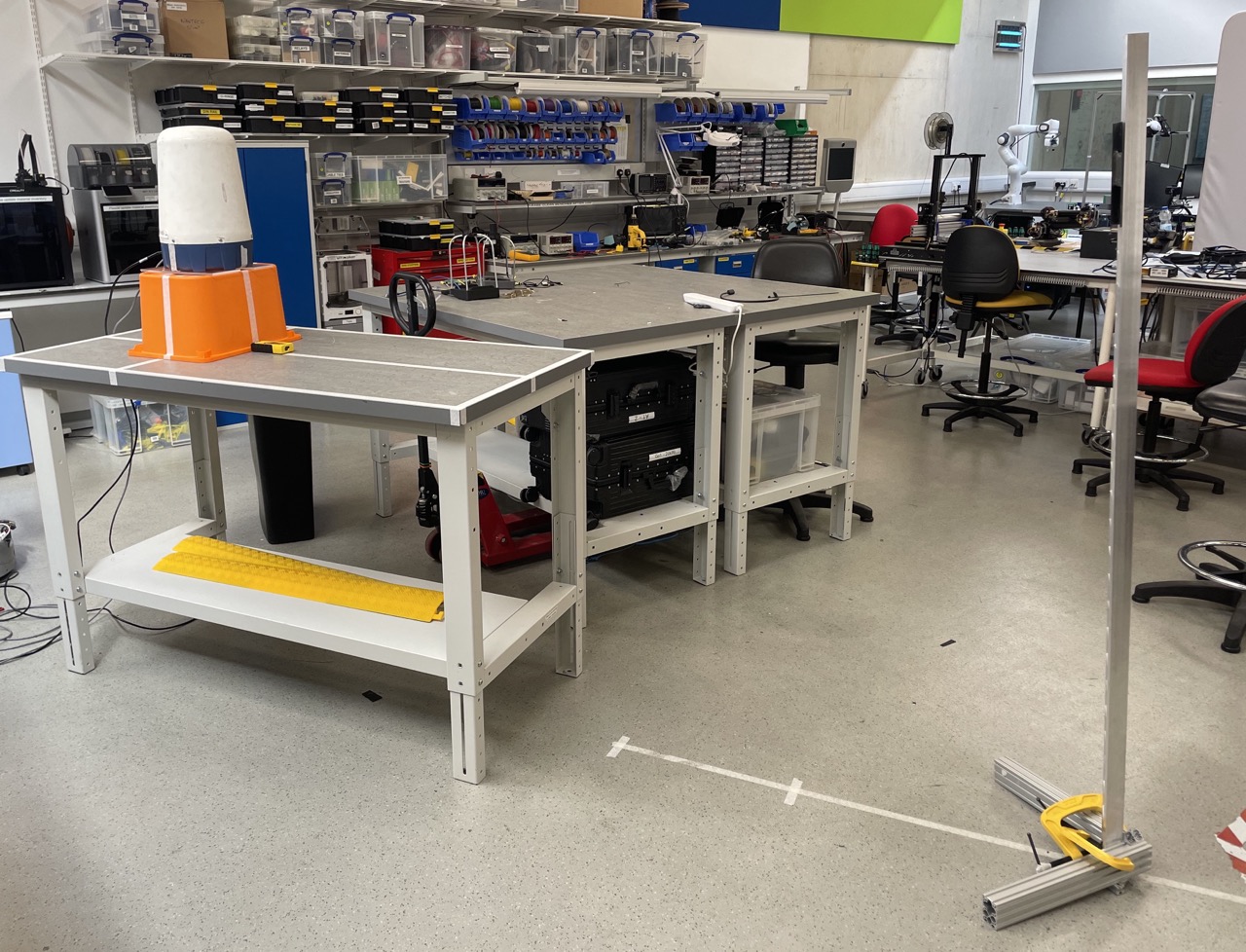}\hfill
    \includegraphics[height=0.37\linewidth]{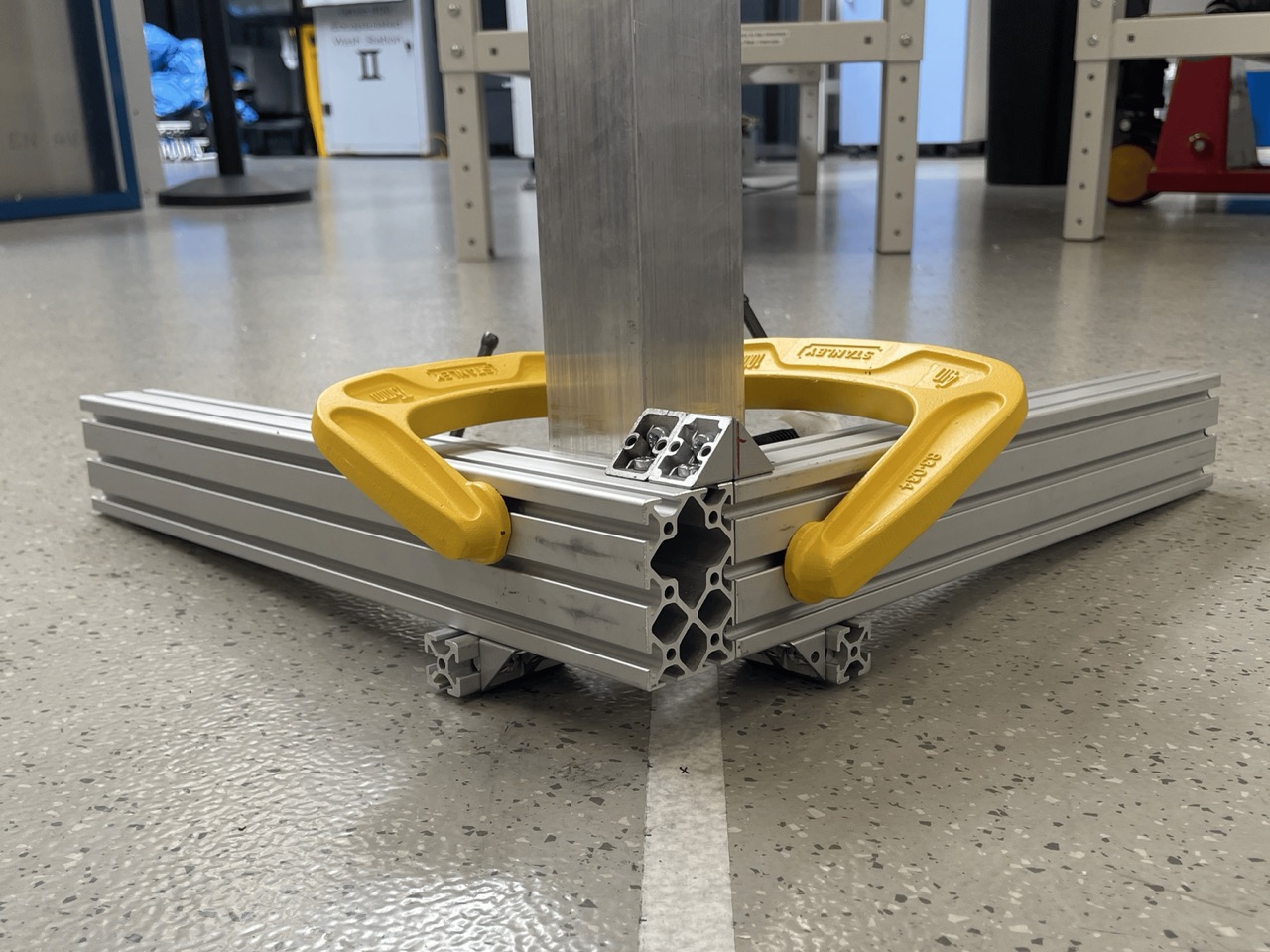}

    \caption{Retroreflective pole used to validate the method: (top row) overview, retroreflector, carbon fibre attenuator; (bottom row) real-world setup and adjustable base .}
    \label{fig:vertical_horizontal}
    \vspace{-10pt}
\end{figure}

The real-world configuration mirrored the simulation but in a static arrangement: a table emulated the chassis shadow, with the Navtech CTS350-X mounted on top and the test pole positioned opposite (\cref{fig:vertical_horizontal}). The radar was positioned to yield $\theta=26^\circ$, with pole alignment marked for repeatability. Distances were measured with a laser rangefinder.

\noindent\textbf{Pole Design.} Initial cardboard and diffuse reflector designs produced weak or inconsistent returns, so a retroreflective pole was built from aluminium profiles with stacked corner reflectors to ensure strong responses across orientations. The final design used an L-profile backbone with aluminium corner reflectors at 6\,cm intervals (\cref{fig:vertical_horizontal}). Carbon fibre sheets at radar height attenuated reflections to prevent saturation, and the base included an adjustable brace to set positive and negative inclinations.

\section{Results}
\label{sec:Results}
\subsection{Simulation Results}
\begin{table*}[!t]
    \centering
    \begin{tabular}{c|cc|cc|cc|cc|cc|cc}
        \hline
        \multirow{2}{*}{Angle}
        & \multicolumn{2}{c|}{Ideal}
        & \multicolumn{2}{c|}{Real res. (raw)}
        & \multicolumn{2}{c|}{Real res. (corrected)}
        & \multicolumn{2}{c|}{Emission model}
        & \multicolumn{2}{c|}{10\,cm pole}
        & \multicolumn{2}{c}{20\,cm pole} \\
        \cline{2-13}
        & Mean & Std & Mean & Std & Mean & Std & Mean & Std & Mean & Std & Mean & Std \\
        \hline
        $-10^\circ$ &  0.034 & 0.105 & 2.011 & 0.907 & 0.044 & 0.831 & 0.078 & 0.843 & 1.777 & 0.964 & 3.765 & 1.234 \\
        $0^\circ$   & -0.009 & 0.107 & 1.371 & 0.593 & 0.035 & 0.543 & 0.048 & 0.513 & 1.244 & 0.626 & 2.720 & 0.752 \\
        $+10^\circ$ & -0.057 & 0.099 & 1.143 & 0.490 & -0.047 & 0.470 & 0.619 & 0.546 & 1.808 & 0.642 & 3.186 & 0.726 \\
        \hline
    \end{tabular}
    \caption{Mean error and standard deviation (degrees) for reconstructed inclination under six simulation setups. 
    \textbf{Ideal}: 1\,cm pole, ideal emission ($0^\circ$ to $-35^\circ$), high resolution (0.002\,m/pixel). 
    \textbf{Real res. (raw)}: same as Ideal but using real radar resolution (0.0438\,m/pixel). 
    \textbf{Real res. (corrected)}: same as Real res. (raw) with half-bin correction.
    \textbf{Emission model}: same as Real res. (corrected) but with measured emission bounds ($+3^\circ$ to $-35^\circ$). 
    \textbf{10\,cm pole}: same as Emission model but with 10\,cm pole. 
    \textbf{20\,cm pole}: same as Emission model but with 20\,cm pole. }
    \label{tab:sim_results_all}    
\end{table*}

We first evaluate the method in simulation, which enables systematic testing under controlled conditions. While simulated scans differ from real radar -- most notably lacking the smudged returns observed in practice -- they validate the geometry analysis and provide a clear view of how resolution, emission profile, and pole diameter influence performance. Summary statistics are reported in \cref{tab:sim_results_all}.

\begin{figure}[!htbp]
    \centering
    \includegraphics[width=0.48\linewidth]{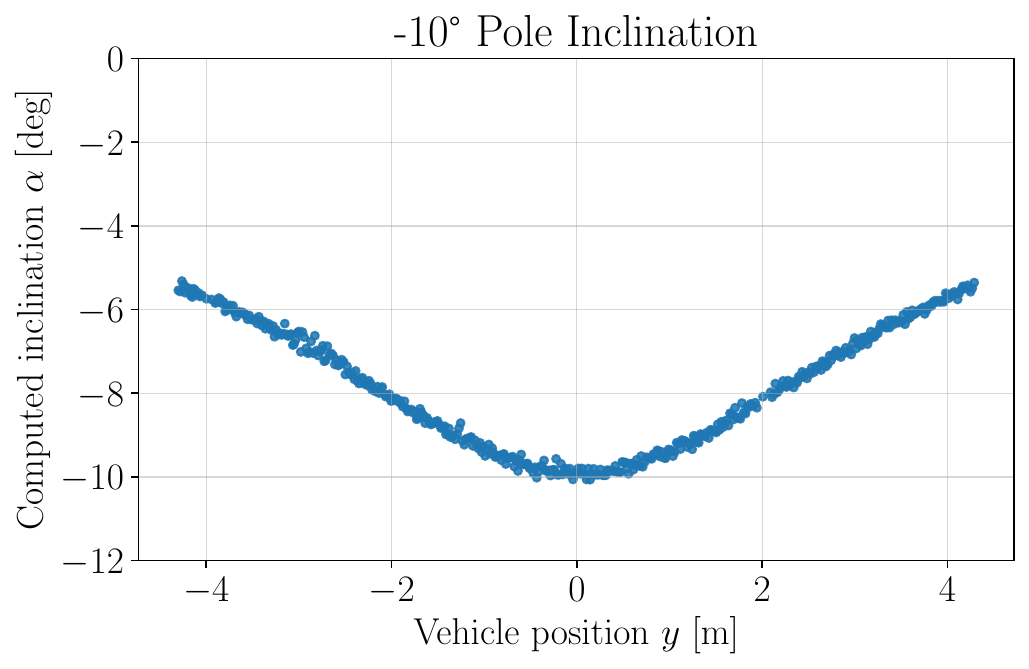}
    \includegraphics[width=0.48\linewidth]{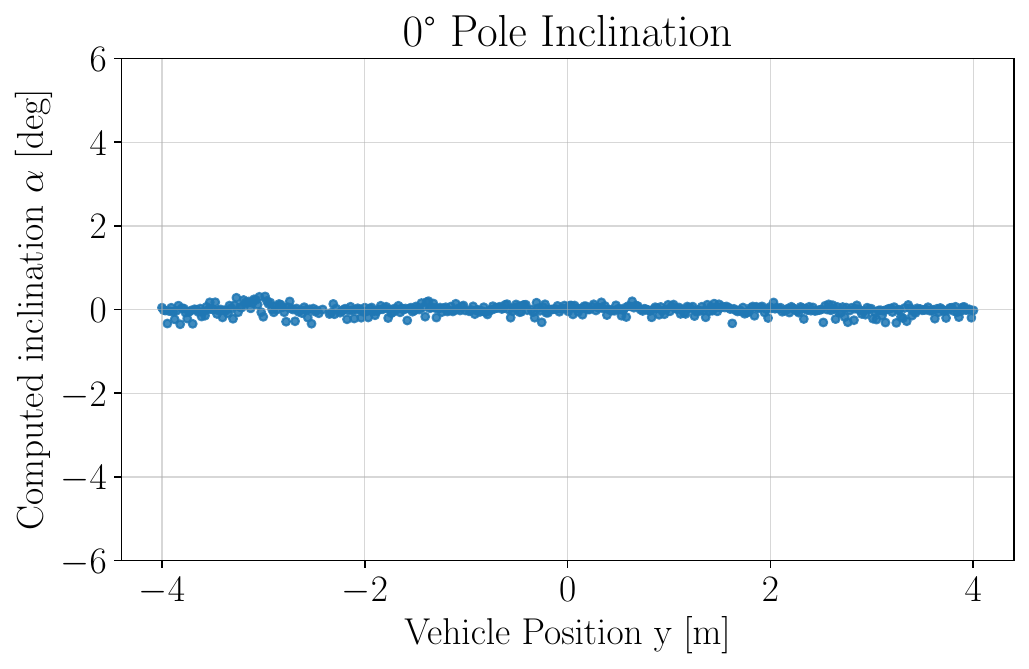}
    \includegraphics[width=0.48\linewidth]{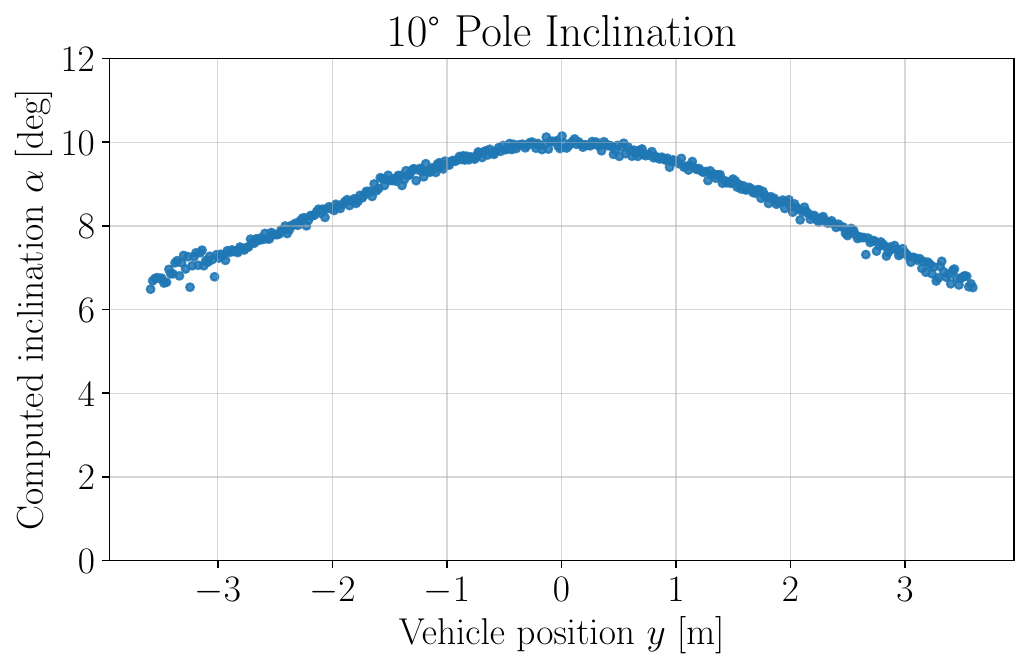}
    \caption{Recovered inclination curves under the ideal setup.}
    \label{fig:case1_plots}
\end{figure}

\noindent\textbf{Results for an Ideal Setup.} Inclinations at $-10^\circ$, $0^\circ$, and $+10^\circ$ are recovered with mean errors below $0.1^\circ$ and standard deviations around $0.1^\circ$ (\cref{tab:sim_results_all}, ``Ideal''), showing that under ideal conditions the method recovers orientation perfectly. We see that the recovered curves for $-10^\circ$ and $+10^\circ$ in \cref{fig:case1_plots} correspond directly to the analytical ground-truth in Fig.~\ref{fig:alpha_graphs_circle}, with $\phi=90^\circ$ and $\phi=270^\circ$ respectively, confirming the validity of the approach.

\begin{figure}[!htbp]
    \centering
    \includegraphics[width=0.48\linewidth]{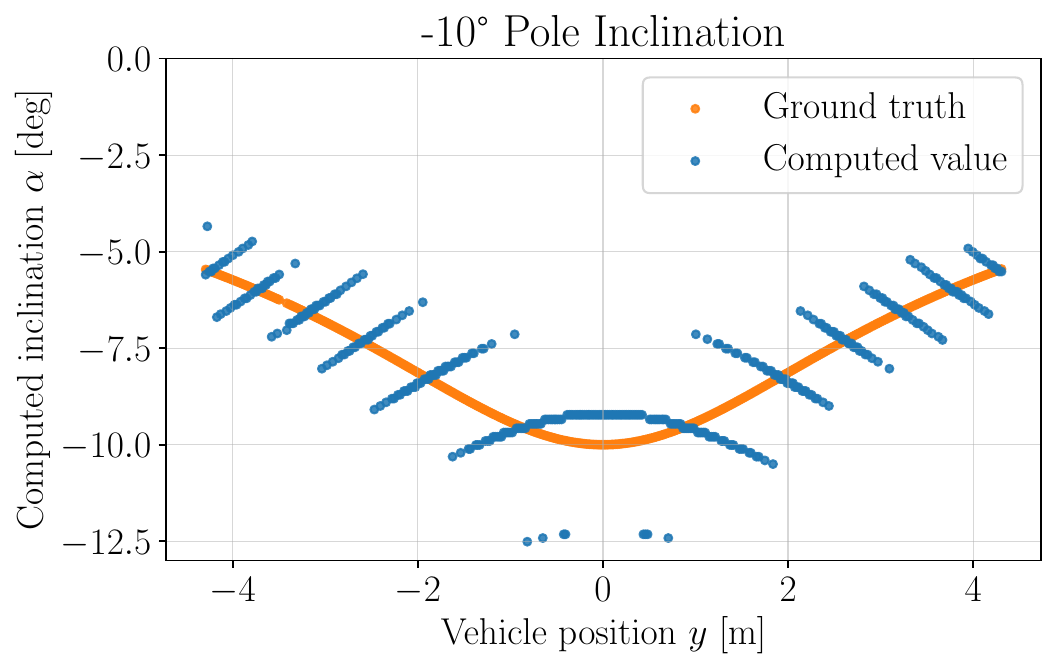}\hfill
    \includegraphics[width=0.48\linewidth]{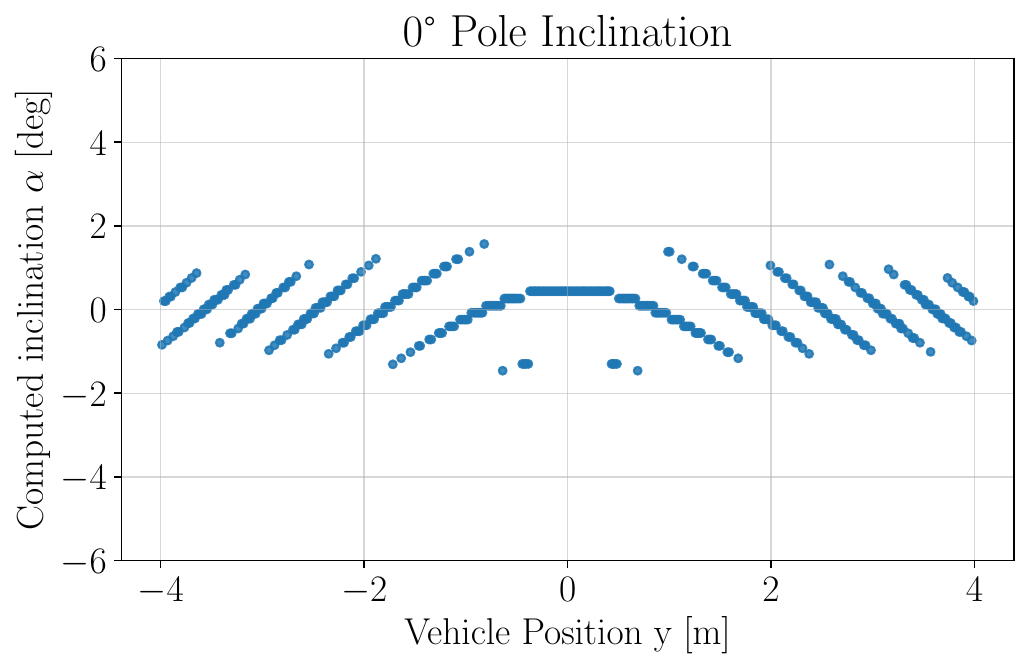}\hfill
    \includegraphics[width=0.48\linewidth]{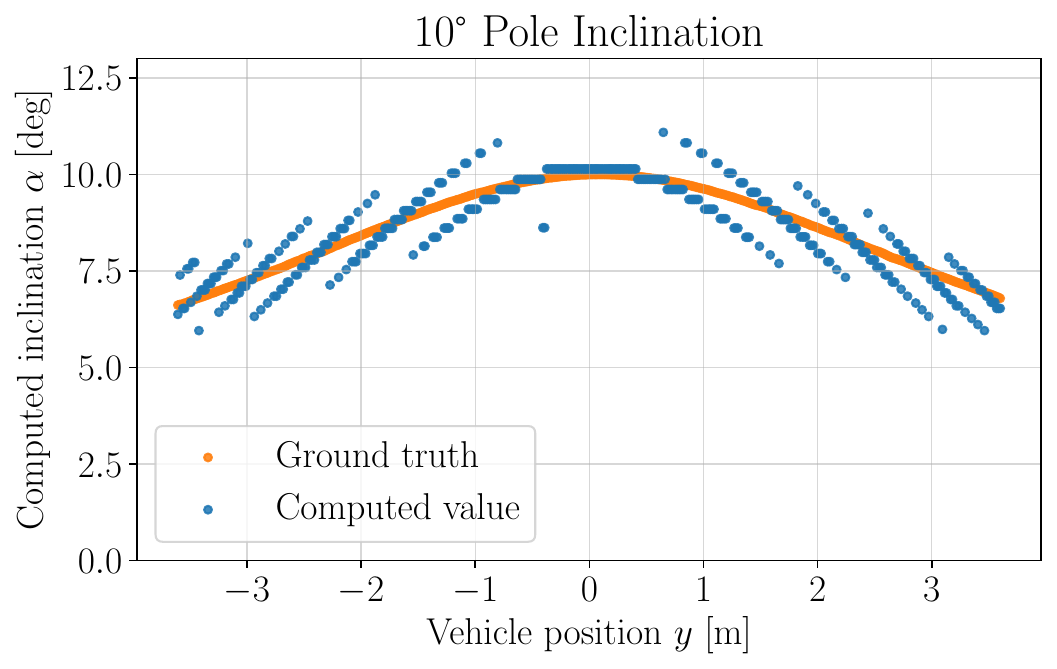}
    \caption{Quantisation with correction: stepped $\alpha(y)$ curves.}
    \label{fig:case2_plots}
\end{figure}

\noindent\textbf{Quantisation Effects.} Because pixels are marked occupied as soon as they are partially filled, $r_1$ is underestimated (front of the first bin) and $r_2$ overestimated (back of the last bin). This shifts the geometry forward and yields a consistent positive bias of 1-2$^\circ$ (\cref{tab:sim_results_all}, ``Real res. (raw)''). A simple half-pixel correction, $\hat r_1=r_1+r/2$, $\hat r_2=r_2-r/2$, removes the bias. After correction, mean errors return close to the ideal case, though curves show a stepped profile due to discretisation (\cref{fig:case2_plots} and \cref{tab:sim_results_all}, ``Real res. (corrected)'').

\noindent\textbf{Emission Model.} Adjusting the emission range to match the CTS350-X leaves performance essentially unchanged, except for a larger bias ($\approx 0.6^\circ$) at $+10^\circ$ (\cref{tab:sim_results_all}, ``Emission model''). The shift arises because only positive tilts form $r_1$ on the upper emission boundary: raising this boundary from the horizontal to $+3^\circ$ shortens $r_1$, so the reconstruction interprets the pole as tilted further forward. For $\alpha=0^\circ$ the effect is negligible, and for $\alpha<0$ it vanishes entirely.

\noindent\textbf{Finite Pole Diameter.} Finite thickness shifts $r_2$ outward: the nearest return comes from the front surface, while the shadow extends to the half-width. This creates a systematic positive bias that grows with diameter, from $\sim1.5^\circ$ at 10\,cm to 2--4$^\circ$ at 20\,cm (Table~\ref{tab:sim_results_all}). Subtracting the radius from $r_2$ would correct it, but this requires prior knowledge of object size.

\noindent\textbf{Simulation takeaway.} Accuracy is essentially perfect under ideal conditions. With quantisation and the real emission model (after correction), errors remain below $0.7^\circ$. In the most challenging setting (e.g., a 20\,cm pole), errors rise to $2$--$4^\circ$ while variance remains similar to other situations.

\subsection{Real-World Results}

\noindent\textbf{Calibration of $d$.}
The elongation bias $d$ was estimated from 110 measurements at 1~cm intervals between 1.75--2.3\,m, defined as the difference between measured $r_2$ and its theoretical true value. A slight negative correlation with distance was observed, but within this range $d$ can be treated as constant (mean 80.6\,cm, std.~6.1\,cm). For the inclination experiments (2.2--2.6\,m), which overlapped by only 15\,cm, the regression fit was extrapolated and its midpoint at 2.38\,m taken to give $d=74.1$\,cm. Segmentation here and in subsequent experiments used 16 frames, approaching a faster real-world deployment.

\begin{figure}[!htbp]
    \centering
    \includegraphics[width=0.85\linewidth]{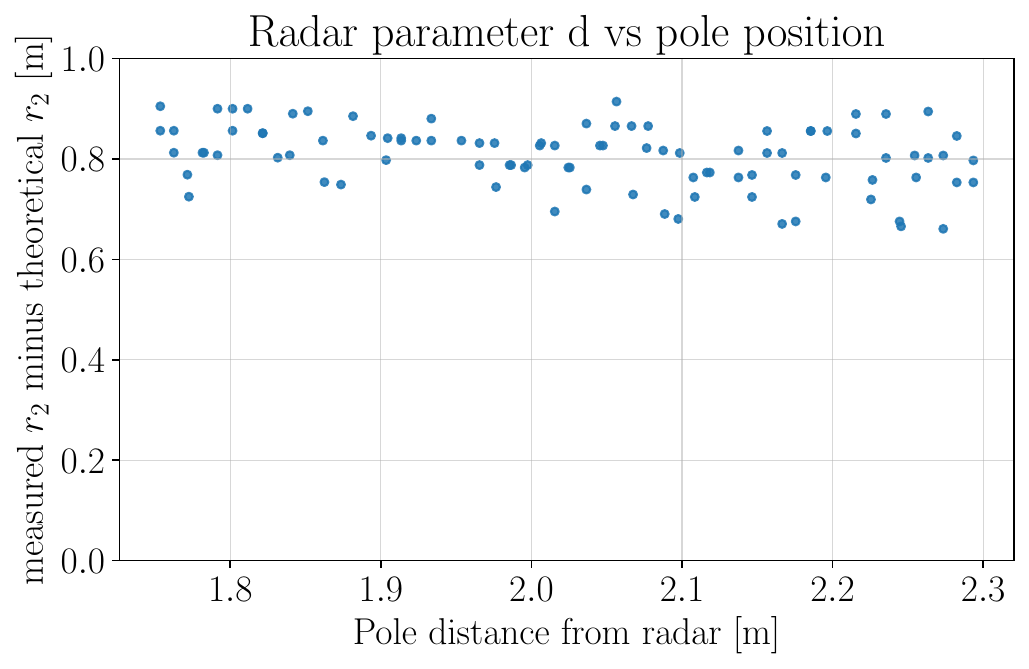}
    \caption{Calibration of elongation bias $d$ from $r_2-r_1$ measurements.}
    \label{fig:calibration_d}
\end{figure}

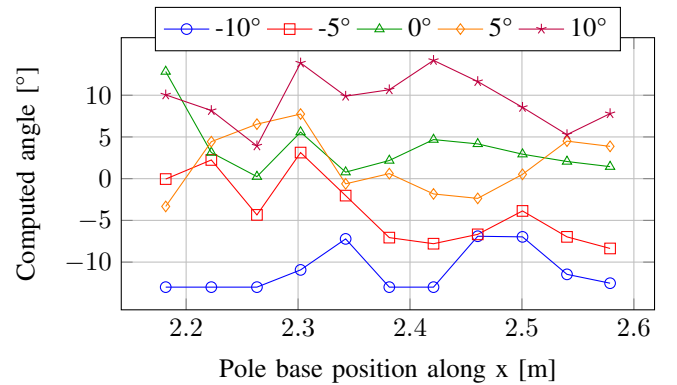
\begin{figure}[!htbp]
    \centering
    \begin{tikzpicture}
        \begin{axis}[
            width=\columnwidth,   
            height =0.6\columnwidth,
            xlabel={Pole base position along x [m]},
            ylabel={Computed angle [°]},
            ytick={-10,-5,0,5,10},
            grid=both,
            legend style={
                at={(0.5,0.95)}, 
                anchor=south,
                legend columns=5
            },
            mark options={solid},
        ]
        
        \addplot[color=blue, mark=o] coordinates {
        (2.18175,-13) (2.2225,-13) (2.26325,-13) (2.30225,-10.9347169) (2.3425,-7.218800143)
        (2.3815,-13) (2.421,-13) (2.4605,-6.901758754) (2.5005,-6.975379723) (2.54025,-11.46779098) (2.57875,-12.52759857)
        };
        \addlegendentry{-10°}
        
        \addplot[color=red, mark=square] coordinates {
        (2.18175,-0.046980751) (2.2225,2.241906027) (2.26325,-4.343521091) (2.30225,3.126480302) (2.3425,-2.030733551)
        (2.3815,-7.06553366) (2.421,-7.795946048) (2.4605,-6.675659139) (2.5005,-3.868026576) (2.54025,-6.975379723) (2.57875,-8.364400155)
        };
        \addlegendentry{-5°}
        
        \addplot[color=green!60!black, mark=triangle] coordinates {
        (2.18175,12.83929428) (2.2225,3.128559791) (2.26325,0.244244991) (2.30225,5.580700483) (2.3425,0.780243283)
        (2.3815,2.177628156) (2.421,4.685472963) (2.4605,4.166466486) (2.5005,2.933312612) (2.54025,2.057021664) (2.57875,1.440475916)
        };
        \addlegendentry{0°}
        
        \addplot[color=orange, mark=diamond] coordinates {
        (2.18175,-3.328936833) (2.2225,4.433920562) (2.26325,6.533535389) (2.30225,7.744537746) (2.3425,-0.618807943)
        (2.3815,0.604677165) (2.421,-1.831575884) (2.4605,-2.363970715) (2.5005,0.518975462) (2.54025,4.500228718) (2.57875,3.870144107)
        };
        \addlegendentry{5°}
        
        \addplot[color=purple, mark=star] coordinates {
        (2.18175,10.04170335) (2.2225,8.169674799) (2.26325,3.930496539) (2.30225,13.8660623) (2.3425,9.88638401)
        (2.3815,10.65572026) (2.421,14.1777207) (2.4605,11.64267506) (2.5005,8.548519798) (2.54025,5.277074781) (2.57875,7.797190575)
        };
        \addlegendentry{10°}
        
        \end{axis}
        \end{tikzpicture}
    \caption{Experimental results for pole inclinations: reconstructed angles across all positions.}
    \label{fig:inclination_results}
    \vspace{-15pt}
\end{figure}

\noindent\textbf{Inclination Experiments.}
The pole was placed at 11 floor marks (4\,cm spacing), with four repetitions each at inclinations of $-10^\circ$, $-5^\circ$, $0^\circ$, $+5^\circ$, and $+10^\circ$. $r_1$ and $r_2$ values were averaged and mapped to inclination. Results are shown in \cref{fig:inclination_results} and \cref{tab:inclination_results}. Inclinations of $-10^\circ$, $-5^\circ$, and $+10^\circ$ are recovered close to ground truth, while $0^\circ$ and $+5^\circ$ show larger shifts. Standard deviations are 3–4$^\circ$, reflecting radar variability not suppressed by averaging 4 frames. Overall, inclination can be recovered within $\approx$3–4$^\circ$, though different effects limit consistency across poses.

\begin{table}[!htbp]
    \centering
    \begin{tabular}{c c c}
        \toprule
        Inclination ($^\circ$) & Mean ($^\circ$) & Std. Dev. ($^\circ$) \\
        \midrule
        $-10$ & $-11.00$ & $2.52$ \\
        $-5$  & $-3.80$  & $3.91$ \\
        $0$   & $+3.64$  & $3.30$ \\
        $+5$  & $+1.82$  & $3.59$ \\
        $+10$ & $+9.45$  & $3.04$ \\
        \bottomrule
    \end{tabular}
    \caption{Inclination reconstruction in real experiments.}
    \label{tab:inclination_results}
\end{table}

\section{Discussion}
\label{sec:Discussion}

\noindent\textbf{Overall performance.} 
Simulation confirmed that pole inclination can be recovered with near-ideal accuracy. Under perfect conditions, errors were $<0.1^\circ$, and after quantisation correction they remained below $0.7^\circ$, even with realistic emission models. Real experiments, however, showed higher variability: inclinations were recovered within $\approx$3--4$^\circ$, with segmentation noise and reflection-dependent variation dominating over quantisation or emission effects. This establishes a clear simulation--reality gap: while simulations isolate geometric factors, real data are constrained by signal strength, segmentation stability, and sensitivity to the calibration of the elongation bias $d$. While an error of $5^\circ$ is significant, distinguishing between an upright pole ($0^\circ$) and one leaning dangerously into the vehicle's path ($>10^\circ$) might prove sufficient to assess traversability in cluttered or unstructured environments, showing potential for future radar-based navigation systems.

\noindent\textbf{Experimental limitations.} The manufactured retroreflective pole was critical to ensure consistent $r_2$ detections, highlighting the challenge of applying the method to natural objects. Carbon-fibre attenuation successfully prevented saturation from the metallic backbone without corrupting $r_2$. The static setup simplified acquisition but omitted temporal return dynamics, which simulations suggested could offer additional constraints for identification. These design choices, though limiting generality, were necessary to demonstrate feasibility in controlled conditions. These limitations could be overcome in the future to make our method more applicable with more sensitive radar sensors and improved segmentation, potentially using more powerful deep learning methods to detect more hidden signals.

\noindent\textbf{Angle-dependent effects.} Errors were not uniform across inclinations. At $-10^\circ$, mean errors were artificially small due to clipping at the theoretical $-13^\circ$ limit. At larger positive angles, variance decreased, consistent with the model prediction that robustness improves as sensitivity to noise decreases with inclination. Simulation-predicted biases from a $+3^\circ$ emission shift and from finite pole diameter were overshadowed by stochastic effects in real-world data. Thus, while the theoretical model captures secondary effects, practical performance is governed by segmentation and reflection variability.

\noindent\textbf{Elongation bias $d$.} Calibration revealed a weak negative correlation of $d$ with distance, but it could be treated as constant over the 40\,cm calibration band. Inclination experiments required extrapolating $d$ to 2.2--2.6\,m, overlapping by only 15\,cm with the calibration range, introducing an additional $\pm$2\,cm uncertainty. Furthermore, $d$ stabilised only when $\geq25$ frames were averaged; using fewer frames, closer to a realistic deployment scenario, required calibration and test data to match in frame count. These results suggest practical deployment remains challenging and techniques more suitable for adaptation to unseen objects should be explored in future works. For example, $d$ could be modelled as strictly range-, object- and frame-dependent, or alternative non-calibrated estimation techniques (e.g. learning-based) could be used to reduce this significant source of error without relying on idealized assumptions.

\noindent\textbf{Noise Sensitivity and Design Considerations.}
Let $\rho=r_2/r_1$ and $D(\rho,\theta)=\rho^2-2\rho\cos\theta+1>0$. For small perturbations,
\[
\delta\alpha \approx \tfrac{\partial\alpha}{\partial \rho}\,\delta \rho + \tfrac{\partial\alpha}{\partial\theta}\,\delta\theta,
\qquad
\]
\[
\mathrm{Var}(\alpha)\approx\Big(\tfrac{\partial\alpha}{\partial \rho}\Big)^2\mathrm{Var}(\rho)+\Big(\tfrac{\partial\alpha}{\partial\theta}\Big)^2\mathrm{Var}(\theta).
\]
For $\alpha\ge0$ ($\rho\ge1/\cos\theta$):
\[
\frac{\partial\alpha}{\partial \rho}=\frac{\sin\theta}{D(\rho,\theta)},\qquad
\frac{\partial\alpha}{\partial\theta}=-\,\frac{\rho(\rho-\cos\theta)}{D(\rho,\theta)}\le -1.
\]
For $\alpha<0$ ($\rho\in[1/\cos(\theta/2),\,1/\cos\theta)$):
\[
\frac{\partial\alpha}{\partial \rho}=\frac{1}{\rho\sqrt{\rho^2-1}},\qquad
\frac{\partial\alpha}{\partial\theta}=-1.
\]

\noindent\textit{Implications.} Sensitivity to $\theta$ is always at least one, so calibration errors directly propagate to $\alpha$. Sensitivity to $\rho$ decreases monotonically with $\rho$ and is bounded by $f_1(\theta)=\cos^2\theta/\sin\theta$ for $\alpha\ge0$ and $f_2(\theta)=\cos^2(\theta/2)/\sin(\theta/2)$ for $\alpha<0$. Both bounds decrease with $\theta$, meaning larger opening angles flatten the $\alpha(\rho)$ curve (\cref{fig:alpha_vs_ratio}) and improve robustness to noise in $\rho$. Negative inclinations are inherently more stable to $\theta$ calibration errors but sensitive to noise in $\rho$, while positive inclinations show the opposite trend.

\def\thetaA{10}   
\def\thetaB{25}
\def\thetaC{40}
\def\thetaD{55}

\def\colA{blue}
\def\colB{red}
\def\colC{green!60!black}
\def\colD{orange}
\begin{figure}[hbt]
\centering
\begin{tikzpicture}
\begin{axis}[
    width=\columnwidth,   
    height =0.5\columnwidth,
    xlabel={$r_{2}/r_{1}$},
    ylabel={$\alpha$ [deg]},
    grid=both,
    legend style={at={(0.97,0.03)},anchor=south east},
    cycle list={}, 
    ymin=-40, ymax=90 
]

\addplot[domain={1/cos(\thetaA/2)}:{1/cos(\thetaA)},samples=240,very thick,\colA]
  {acos(1/x) - \thetaA};
\addplot[domain={1/cos(\thetaA)}:10,samples=240,very thick,\colA,forget plot]
  {atan((x*cos(\thetaA) - 1)/(x*sin(\thetaA)))};
\addlegendentry{$\theta=\thetaA^\circ$}

\addplot[domain={1/cos(\thetaB/2)}:{1/cos(\thetaB)},samples=240,very thick,\colB]
  {acos(1/x) - \thetaB};
\addplot[domain={1/cos(\thetaB)}:10,samples=240,very thick,\colB,forget plot]
  {atan((x*cos(\thetaB) - 1)/(x*sin(\thetaB)))};
\addlegendentry{$\theta=\thetaB^\circ$}

\addplot[domain={1/cos(\thetaC/2)}:{1/cos(\thetaC)},samples=240,very thick,\colC]
  {acos(1/x) - \thetaC};
\addplot[domain={1/cos(\thetaC)}:10,samples=240,very thick,\colC,forget plot]
  {atan((x*cos(\thetaC) - 1)/(x*sin(\thetaC)))};
\addlegendentry{$\theta=\thetaC^\circ$}

\addplot[domain={1/cos(\thetaD/2)}:{1/cos(\thetaD)},samples=240,very thick,\colD]
  {acos(1/x) - \thetaD};
\addplot[domain={1/cos(\thetaD)}:10,samples=240,very thick,\colD,forget plot]
  {atan((x*cos(\thetaD) - 1)/(x*sin(\thetaD)))};
\addlegendentry{$\theta=\thetaD^\circ$}

\addplot[dashed,black,thick,domain=1:10]{0};

\end{axis}
\end{tikzpicture}
\caption{Dependence of $\alpha$ on $r_{2}/r_{1}$ for representative $\theta$. Larger $\theta$ values flatten the curve, reducing sensitivity to noise in $x$.}
\label{fig:alpha_vs_ratio}
\end{figure}
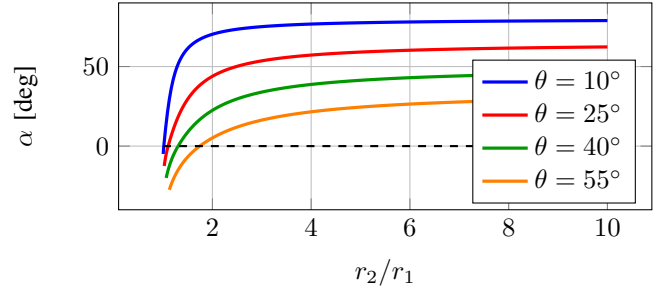

\noindent\textit{Design recommendations.} Increasing $\theta$ improves robustness but shortens the shadow cone and limits range. Calibration of $\theta$ is therefore critical, while the ratio $\rho$ governs noise sensitivity: values near $1$ (negative $\alpha$) are less stable, whereas larger $\rho$ improves reliability. This trade-off between robustness and operational reach suggests two strategies: (i) an active occluding mask to adapt $\theta$ for long- vs.\ short-range conditions, or (ii) adjusting radar height to recover reach when using larger $\theta$ values.

\section{Conclusion}
We introduce a geometric method to estimate the inclination of slender vertical objects using only 2D FMCW radar. By exploiting the chassis-induced shadow as a stable reference, the method turns an overlooked artefact into a structural cue. We present a closed-form mapping from $(r_1,r_2,\theta)$ to inclination, which we have validated in simulation, and tested with a Navtech CTS350-X. Our results show near-ideal accuracy in controlled simulation ($<0.7^\circ$ mean error) and practical feasibility in real data with errors of $\approx$3--4$^\circ$.

Overall, this work demonstrates that mechanically rotating radar can provide richer geometric cues than previously assumed, turning an overlooked property into useful spatial information. By formalising and exploiting the chassis shadow, we hope to inspire future work in which radar shadows can provide a viable source of information for 3D geometry-aware perception in robotics.
While showing potential and highlighting new ways to utilise information from radar scans, our approach still has clear limitations and serves primarily as a proof-of-concept, constrained by idealised assumptions, such as a flat ground plane and a perfectly slender target.
Future work should therefore prioritise testing on generic objects without specific reflection structures and refining signal processing to bridge the sim-to-real gap under dynamic driving conditions, extend to adaptively handle uneven ground or correct for varying object thickness to broaden applicability.


\bibliographystyle{IEEEtran}
\bibliography{biblio}

@inproceedings{BijelicCVPR2020_SeeingThroughFog,
  author    = {Bijeli{\'c}, Mario and Gruber, Tobias and Mannan, Fahim and Kraus, Florian and Ritter, Werner and Dietmayer, Klaus and Heide, Felix},
  title     = {Seeing Through Fog Without Seeing Fog: Deep Multimodal Sensor Fusion in Unseen Adverse Weather},
  booktitle = {Proc. IEEE/CVF Conf. on Computer Vision and Pattern Recognition (CVPR)},
  year      = {2020},
  pages     = {11682--11692},
  doi       = {10.1109/CVPR42600.2020.01170}
}

@INPROCEEDINGS{cen2019,
  author={Cen, Sarah H. and Newman, Paul},
  booktitle={2019 International Conference on Robotics and Automation (ICRA)}, 
  title={Radar-only ego-motion estimation in difficult settings via graph matching}, 
  year={2019},
  volume={},
  number={},
  pages={298-304},
  doi={10.1109/ICRA.2019.8793990}}

@INPROCEEDINGS{cfear,
  author={Adolfsson, Daniel and Magnusson, Martin and Alhashimi, Anas and Lilienthal, Achim J. and Andreasson, Henrik},
  booktitle={2021 IEEE/RSJ International Conference on Intelligent Robots and Systems (IROS)}, 
  title={CFEAR Radarodometry - Conservative Filtering for Efficient and Accurate Radar Odometry}, 
  year={2021},
  pages={5462-5469},
  doi={10.1109/IROS51168.2021.9636253}}

@article{burnett2021radar,
  title={Radar odometry combining probabilistic estimation and unsupervised feature learning},
  author={Burnett, Keenan and Yoon, David J and Schoellig, Angela P and Barfoot, Timothy D},
  journal={arXiv preprint arXiv:2105.14152},
  year={2021}
}

@INPROCEEDINGS{radarslam,
  author={Hong, Ziyang and Petillot, Yvan and Wang, Sen},
  booktitle={2020 IEEE/RSJ International Conference on Intelligent Robots and Systems (IROS)}, 
  title={RadarSLAM: Radar based Large-Scale SLAM in All Weathers}, 
  year={2020},
  volume={},
  number={},
  pages={5164-5170},
  doi={10.1109/IROS45743.2020.9341287}}

@article{tang2023point,
  title={Point-based metric and topological localisation between lidar and overhead imagery},
  author={Tang, Tim Yuqing and De Martini, Daniele and Newman, Paul},
  journal={Autonomous Robots},
  volume={47},
  number={5},
  pages={595--615},
  year={2023},
  publisher={Springer}
}

@article{tang2021self,
  title={Self-supervised learning for using overhead imagery as maps in outdoor range sensor localization},
  author={Tang, Tim Y and De Martini, Daniele and Wu, Shangzhe and Newman, Paul},
  journal={The International Journal of Robotics Research},
  volume={40},
  number={12-14},
  pages={1488--1509},
  year={2021},
  publisher={SAGE Publications Sage UK: London, England}
}

@inproceedings{HahnerICCV2021_FogSimLiDAR,
  author    = {Hahner, Martin and Sakaridis, Christos and Dai, Dengxin and Van Gool, Luc},
  title     = {Fog Simulation on Real {LiDAR} Point Clouds for 3D Object Detection in Adverse Weather},
  booktitle = {Proc. IEEE/CVF Int. Conf. on Computer Vision (ICCV)},
  year      = {2021},
  pages     = {15283--15292},
  doi       = {10.1109/ICCV48922.2021.01501}
}

@inproceedings{HahnerCVPR2022_SnowLiDAR,
  author    = {Hahner, Martin and Sakaridis, Christos and Bijeli{\'c}, Mario and Heide, Felix and Yu, Fisher and Dai, Dengxin and Van Gool, Luc},
  title     = {{LiDAR} Snowfall Simulation for Robust 3D Object Detection},
  booktitle = {Proc. IEEE/CVF Conf. on Computer Vision and Pattern Recognition (CVPR)},
  year      = {2022},
  pages     = {16364--16374},
  doi       = {10.1109/CVPR52688.2022.01602}
}

@inproceedings{PaekNeurIPS2022_KRadar,
  author    = {Paek, Dong-Hee and Kong, Seung-Hyun and Wijaya, Kevin Tirta},
  title     = {K-{Radar}: 4D Radar Object Detection for Autonomous Driving in Various Weather Conditions},
  booktitle = {NeurIPS Datasets and Benchmarks Track},
  year      = {2022}
}

@inproceedings{CaesarCVPR2020_nuScenes,
  author    = {Caesar, Holger and Bankiti, Varun and Lang, Alex H. and Vora, Sourabh and Liong, Venice Erin and Xu, Qiang and Krishnan, Anush and Pan, Yu and Baldan, Giancarlo and Beijbom, Oscar},
  title     = {nuScenes: A Multimodal Dataset for Autonomous Driving},
  booktitle = {Proc. IEEE/CVF Conf. on Computer Vision and Pattern Recognition (CVPR)},
  year      = {2020},
  pages     = {11618--11628},
  doi       = {10.1109/CVPR42600.2020.01165}
}

@article{AllandSPM2019_InterferenceSurvey,
  author  = {Alland, Stephen and Stark, Wayne E. and Ali, Murtaza and Hegde, Manju V.},
  title   = {Interference in Automotive Radar Systems: Characteristics, Mitigation Techniques, and Current and Future Research},
  journal = {IEEE Signal Processing Magazine},
  year    = {2019},
  volume  = {36},
  number  = {5},
  pages   = {45--59},
  doi     = {10.1109/MSP.2019.2908214}
}

@inproceedings{KrausIROS2021_GhostObjects,
  author    = {Kraus, Florian and Scheiner, Nicolas and Ritter, Werner and Dietmayer, Klaus},
  title     = {The Radar Ghost Dataset -- An Evaluation of Ghost Objects in Automotive Radar Data},
  booktitle = {Proc. IEEE/RSJ Int. Conf. on Intelligent Robots and Systems (IROS)},
  year      = {2021},
  pages     = {8570--8577},
  doi       = {10.1109/IROS51168.2021.9636338}
}

@article{Zhang2023_AdverseWeatherSurvey,
  author  = {Zhang, Yuxiao and Carballo, Alexander and Yang, Hanting and Takeda, Kazuya},
  title   = {Perception and Sensing for Autonomous Vehicles Under Adverse Weather Conditions: A Survey},
  journal = {ISPRS Journal of Photogrammetry and Remote Sensing},
  year    = {2023},
  volume  = {196},
  pages   = {146--177},
  doi     = {10.1016/j.isprsjprs.2022.12.021}
}

@article{Zhou2022_DeepRadarPerceptionSurvey,
  author  = {Zhou, Yi and Liu, Lulu and Zhao, Haocheng and L{\'o}pez-Ben{\'i}tez, Miguel and Yu, Limin and Yue, Yutao},
  title   = {Towards Deep Radar Perception for Autonomous Driving: Datasets, Methods, and Challenges},
  journal = {Sensors},
  year    = {2022},
  volume  = {22},
  number  = {11},
  pages   = {4208},
  doi     = {10.3390/s22114208}
}

@inproceedings{CenICRA2018_PreciseRadarEgomotion,
  author    = {Cen, Sarah H. and Newman, Paul},
  title     = {Precise Ego-Motion Estimation with Millimeter-Wave Radar Under Diverse and Challenging Conditions},
  booktitle = {Proc. IEEE Int. Conf. on Robotics and Automation (ICRA)},
  year      = {2018},
  pages     = {6045--6052},
  doi       = {10.1109/ICRA.2018.8460687}
}

@inproceedings{BarnesCoRL2020_MaskingByMoving,
  author    = {Barnes, Dan and Weston, Rob and Posner, Ingmar},
  title     = {Masking by Moving: Learning Distraction-Free Radar Odometry from Pose Information},
  booktitle = {Proc. Conf. on Robot Learning (CoRL)},
  year      = {2020}
}

@inproceedings{Burnett2021_RadarOdometryRSS,
  author    = {Burnett, Keenan and Yoon, David J. and Schoellig, Angela P. and Barfoot, Timothy D.},
  title     = {Radar Odometry Combining Probabilistic Estimation and Unsupervised Feature Learning},
  booktitle = {Proceedings of Robotics: Science and Systems (RSS)},
  year      = {2021},
  doi       = {10.15607/RSS.2021.XVII.029}
}

@inproceedings{Qiao2025_RadarTeachRepeat,
  author    = {Qiao, Xinyuan and Krawciw, Alexander and Lilge, Sven and Barfoot, Timothy D.},
  title     = {Radar Teach and Repeat: Architecture and Initial Field Testing},
  booktitle = {Proceedings of the IEEE International Conference on Robotics and Automation (ICRA)},
  year      = {2025},
  doi       = {10.1109/ICRA55743.2025.11128412}
}

@inproceedings{Borts2024_RadarFields,
  author    = {Borts, David and Liang, Erich and Brödermann, Tim and Ramazzina, Andrea and Walz, Stefanie and Palladin, Edoardo and Sun, Jipeng and Brueggemann, David and Sakaridis, Christos and Van Gool, Luc and Bijelić, Mario and Heide, Felix},
  title     = {Radar Fields: Frequency-Space Neural Scene Representations for FMCW Radar},
  booktitle = {ACM SIGGRAPH 2024 Conference Papers},
  year      = {2024},
  address   = {New York, NY, USA},
  doi       = {10.1145/3641519.3657510},
  articleno = {130},
  numpages  = {10},
  publisher = {Association for Computing Machinery},
  isbn      = {9798400705250}
}

@article{Yang2020_VideoSAR_ShadowGMTI,
  author  = {Yang, Xiaqing and Shi, Jun and Zhou, Yuanyuan and Wang, Chen and Hu, Yao and Zhang, Xiaoling and Wei, Shunjun},
  title   = {Ground Moving Target Tracking and Refocusing Using Shadow in Video-{SAR}},
  journal = {Remote Sensing},
  year    = {2020},
  volume  = {12},
  number  = {18},
  pages   = {3083},
  doi     = {10.3390/rs12183083}
}

@article{Sun2022_SAR_BuildingHeight,
  author  = {Sun, Yao and Mou, Lichao and Wang, Yuanyuan and Montazeri, Sina and Zhu, Xiao Xiang},
  title   = {Large-Scale Building Height Retrieval From Single {SAR} Imagery Based on Bounding Box Regression Networks},
  journal = {ISPRS Journal of Photogrammetry and Remote Sensing},
  year    = {2022},
  volume  = {184},
  pages   = {79--95},
  doi     = {10.1016/j.isprsjprs.2021.11.024}
}

@article{Xie2021_SAR_ShadowHeight,
  author  = {Xie, Yakun and Feng, Dejun and Xiong, Sifan and Zhu, Jun and Liu, Yangge},
  title   = {Multi-Scene Building Height Estimation Method Based on Shadow in High Resolution Imagery},
  journal = {Remote Sensing},
  year    = {2021},
  volume  = {13},
  number  = {15},
  pages   = {2862},
  doi     = {10.3390/rs13152862}
}

@inproceedings{BarnesICRA2020_ORRD,
  author    = {Barnes, Dan and Gadd, Matthew and Murcutt, Paul and Newman, Paul and Posner, Ingmar},
  title     = {The Oxford Radar RobotCar Dataset: A Radar Extension to the Oxford RobotCar Dataset},
  booktitle = {Proc. IEEE Int. Conf. on Robotics and Automation (ICRA)},
  year      = {2020},
  pages     = {6433--6438},
  doi       = {10.1109/ICRA40945.2020.9196674}
}

@article{Gadd2024_OORD,
  author    = {Gadd, Matthew and De Martini, Daniele and Bartlett, Oliver and Murcutt, Paul and Towlson, Matt and Widojo, Matthew and Mu{\c{s}}at, Valentina and Robinson, Luke and Panagiotaki, Efimia and Pramatarov, Georgi and K{\"u}hn, Marc Alexander and Marchegiani, Letizia and Newman, Paul and Kunze, Lars},
  title     = {{OORD}: The Oxford Offroad Radar Dataset},
  journal   = {IEEE Transactions on Intelligent Transportation Systems},
  year      = {2024},
  volume    = {25},
  number    = {11},
  pages     = {18779--18790},
  doi       = {10.1109/TITS.2024.3424984},
}

@article{Mock2023_RadaRays,
  author={Mock, Alexander and Magnusson, Martin and Hertzberg, Joachim},
  journal={IEEE Robotics and Automation Letters}, 
  title={RadaRays: Real-Time Simulation of Rotating FMCW Radar for Mobile Robotics via Hardware-Accelerated Ray Tracing}, 
  year={2025},
  volume={10},
  number={3},
  pages={2470-2477},
  doi={10.1109/LRA.2025.3531689}
}

@inproceedings{Dosovitskiy2017_CARLA,
  author    = {Dosovitskiy, Alexey and Ros, German and Codevilla, Felipe and Lopez, Antonio and Koltun, Vladlen},
  title     = {{CARLA}: An Open Urban Driving Simulator},
  booktitle = {Proceedings of the 1st Annual Conference on Robot Learning (CoRL)},
  year      = {2017},
  volume    = {78},
  pages     = {1--16},
  series    = {Proceedings of Machine Learning Research (PMLR)}
}

@inproceedings{radarsplat,
          title={{RadarSplat: Radar Gaussian Splatting for High-Fidelity Data Synthesis and 3D Reconstruction of Autonomous Driving Scenes}},
          author={Kung, Pou-Chun and Harisha, Skanda and Vasudevan, Ram and Eid, Aline and Skinner, Katherine A},
          booktitle={Proceedings of the IEEE/CVF International Conference on Computer Vision},
          pages={27596--27606},
          year={2025}
        }

@INPROCEEDINGS{hou2023automotive,
  author={Hou, Chun-Yu and Wang, Chieh-Chih and Lin, Wen-Chieh},
  booktitle={2023 IEEE/RSJ International Conference on Intelligent Robots and Systems (IROS)}, 
  title={{Automotive Radar Missing Dimension Reconstruction from Motion}}, 
  year={2023},
  volume={},
  number={},
  pages={11226-11232},
  doi={10.1109/IROS55552.2023.10342167}}

\end{document}